\documentclass[review]{elsarticle}
\usepackage{times}
\usepackage{epsfig}
\usepackage{lineno,hyperref}
\usepackage{graphicx}
\usepackage{caption}
\usepackage{amsmath}
\usepackage{amssymb}
\usepackage{url}            
\usepackage{booktabs}       
\usepackage{amsfonts}       
\usepackage{nicefrac}       
\usepackage{microtype}      
\usepackage{multirow}
\usepackage{verbatim}
\usepackage{xcolor}
\usepackage{float}
\usepackage{csquotes}
\usepackage{setspace}
\usepackage[ruled]{algorithm2e}

\newcommand{\XX}{\mathbf{X}}
\newcommand{\yy}{\mathbf{y}}
\newcommand{\zz}{\mathbf{z}}
\newcommand{\pp}{\mathbf{p}}
\newcommand{\real}{\mathbb{R}}

\hypersetup{pdfauthor=author}
\modulolinenumbers[5]

\journal{Journal of Pattern Recognition}

\begin{document}

\begin{frontmatter}

\title{Incremental Multi-Target Domain Adaptation for\\ Object Detection with Efficient Domain Transfer}

\author{Le Thanh Nguyen-Meidine\corref{cor1}\textsuperscript{$a$}}
\ead{le-thanh.nguyen-meidine.1@ens.etsmtl.ca}

\author{Madhu Kiran\textsuperscript{$a$}}
\ead{madhu.kiran.1@ens.etsmtl.ca}

\author{Marco Pedersoli\textsuperscript{$a$}}
\ead{Marco.Pedersoli@etsmtl.ca}

\author{Jose Dolz\textsuperscript{$a$}}
\ead{jose.dolz@etsmtl.ca}

\author{Louis-Antoine Blais-Morin\textsuperscript{$b$}}
\ead{lablaismorin@genetec.com}

\author{Eric Granger\textsuperscript{$a$}}
\ead{eric.granger@etsmtl.ca}

\address[mysecondaryaddress]{Laboratoire d'imagerie, de vision et d'intelligence artificielle (LIVIA) \\ \'Ecole de technologie sup\'erieure, Montreal, Canada.}

\address[mysecondaryaddress]{Genetec Inc., Montreal, Canada.}

\cortext[cor1]{Corresponding author}

\begin{abstract}  \setstretch{1.3}
Recent advances in unsupervised domain adaptation have significantly improved the recognition accuracy of CNNs by alleviating the domain shift between (labeled) source and (unlabeled) target data distributions. While the problem of single-target domain adaptation (STDA) for object detection has recently received much attention, multi-target domain adaptation (MTDA) remains largely unexplored, despite its practical relevance in several real-world applications, such as multi-camera video surveillance. Compared to the STDA problem that may involve large domain shifts between complex source and target distributions, MTDA faces additional challenges, most notably the computational requirements and catastrophic forgetting of previously-learned targets, which can depend on the order of target adaptations. STDA for detection can be applied to MTDA by adapting one model per target, or one common model with a mixture of data from target domains. However, these approaches are either costly or inaccurate. The only state-of-art MTDA method specialized for detection learns targets incrementally, one target at a time, and mitigates the loss of knowledge by using a duplicated detection model for knowledge distillation, which is computationally expensive and does not scale well to many domains. 
%
In this paper, we introduce an efficient approach for incremental learning that generalizes well to multiple target domains. Our MTDA approach is more suitable for real-world applications since it allows updating the detection model incrementally, without storing data from previous-learned target domains, nor retraining when a new target domain becomes available. 
Our approach leverages domain discriminators to train a novel Domain Transfer Module (DTM), which only incurs a modest overhead. The DTM transforms source images according to diverse target domains, allowing the model to access a joint representation of previously-learned target domains, and to effectively limit catastrophic forgetting. Our proposed method -- called MTDA with DTM (MTDA-DTM) -- is compared against state-of-the-art approaches on several MTDA detection benchmarks and Wildtrack, a benchmark for multi-camera pedestrian detection. Results indicate that MTDA-DTM achieves the highest level of detection accuracy across multiple target domains, yet requires significantly fewer computational resources. Our code is available\href{https://github.com/Natlem/M-HTCN}{}\footnote{https://github.com/Natlem/M-HTCN}.
\end{abstract}


\end{frontmatter}

\section{Introduction}

With the advent of deep learning (DL) models such as Faster-CNN \cite{frcnn}, object detection has experienced significant improvements. Despite the high level of accuracy provided by these models in a wide range of benchmark datasets and applications \cite{zhao2019object, efficientdet}, current object detectors still suffer from poor generalization in the presence of domain shift between training and test datasets. For instance, in video surveillance applications, videos are captured over a distributed set of cameras with non-overlapping viewpoints. The shift between the source (e.g., lab setting) and target (e.g., cameras) domains may lead to a significant decline in detection accuracy.  A straightforward solution would involve a supervised fine-tuning of a given model using annotated target domain samples. However, the high cost associated with the collection and annotation of target data may prohibit the adoption and deployment of such models in many practical scenarios. 

\begin{figure*}[!t]
    \centering
    \includegraphics[width=0.8\textwidth]{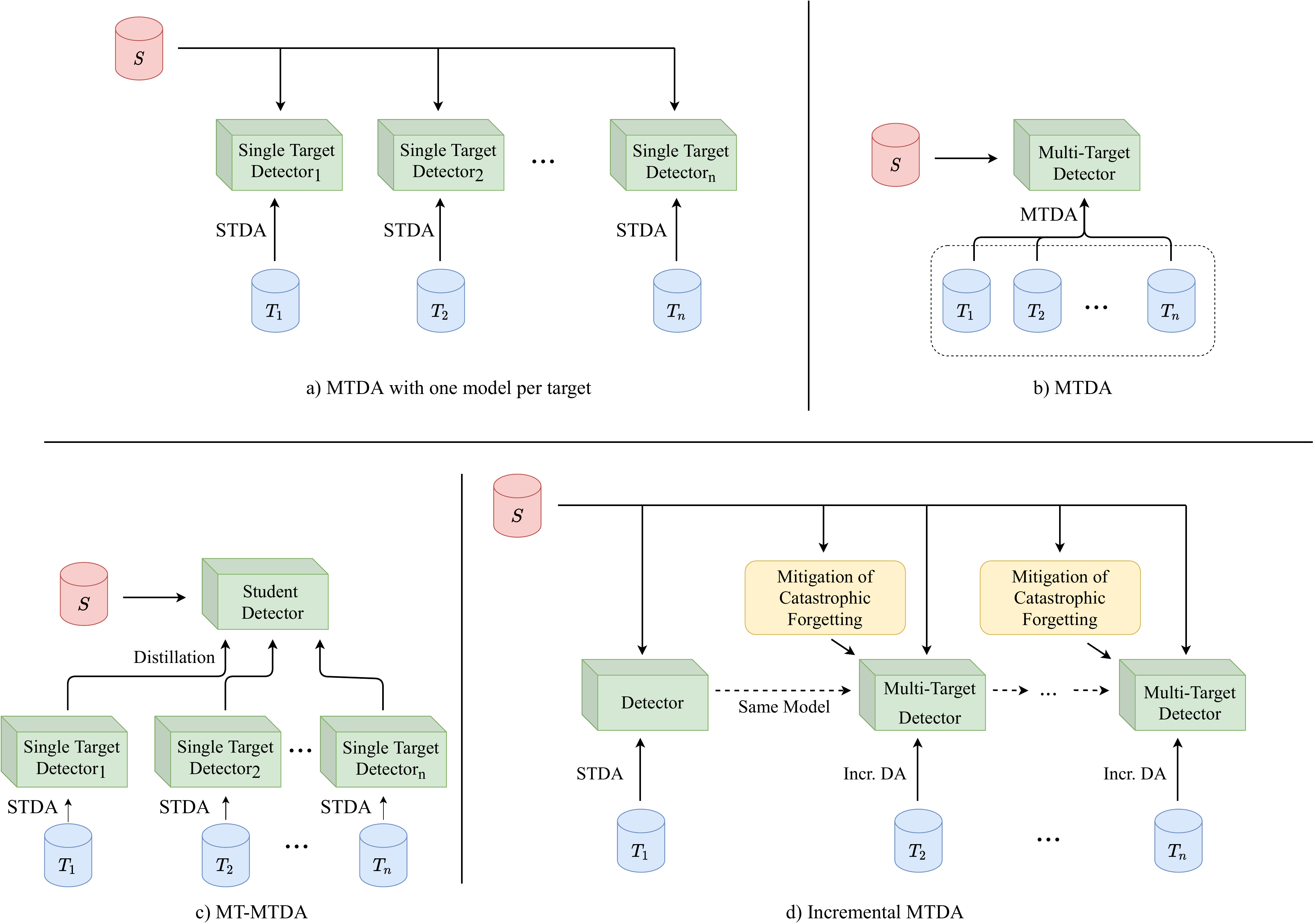}
    \caption{Illustration of different MTDA strategies for object detection:  a) MTDA with a  detection model per target. b) MTDA with one common model trained with a mixture of data from all targets or with separate targets. c) MTDA with one teacher detector per target domain and a common model is obtained by through knowledge distillation d) Incremental MTDA, where a common model is adapted incrementally on one target at time using an additional module (mitigation of catastrophic forgetting, i.e. duplicated model for distillation in \cite{wei2020_inc_mtda}) that limits catastrophic forgetting. ($S$ represents the source domain dataset, and $T_1, T_2, ... T_n$ the $n$ target domain datasets.}
    \label{fig:mtda_scenario}
    \vspace{-3mm}
\end{figure*}

Unsupervised Domain Adaptation (UDA) has been proposed to alleviate the problem of domain shift when labeled target data is not available. Depending on the number of target domains, UDA methods may either address single-target domain adaptation (STDA) or multi-target domain adaptation (MTDA) problems. STDA methods range from learning discriminant domain-invariant features using an adversarial loss \cite{FRCNN_DA, GRL}, to learning mappings between a source and target domain for domain transfer \cite{HTCN, Progressive_DA_FRCNN}, or to using well-known generative models, such as CycleGAN \cite{CycleGAN2017} in combination with previous methods for learning domain-invariant features. Existing STDA techniques can be extended to MTDA by either adapting multiple object detection models, one for each target domain (Fig.\ref{fig:mtda_scenario}~(a)), or by treating multiple target domains as one target domain, and then applying STDA on a mixture of target data. Nevertheless, these approaches are complex because the number of models equals the number of target domains (Fig.\ref{fig:mtda_scenario}~(a)), or generalize poorly on several distinct target domains, particularly when the number of target domains increases as shown in \cite{ mt-mtda, wei2020_inc_mtda}. The poor performance of STDA in the MTDA setting, especially in object detection, is mainly due to the fact that generalizing on multiple target domains as a single domain is difficult since the underlying data distribution of these targets are still very different. This is why specific MTDA techniques need to be developed. In this way, the model can learn multiple underlying data distributions without sacrificing computation cost by using only one common multiple-target model.

Despite its importance for many real-world applications, MTDA remains largely unexplored. For example, given a multi-camera detection problem in video surveillance, where videos are captured using several cameras with different viewpoints, backgrounds, and capture conditions (i.e., target domain), MTDA methods are required so that the object detector may generalize well to multiple target domains. Several techniques for MTDA have been proposed in the context of image classification \cite{BlendMTDA, dada}. Among these, AMEAN \cite{BlendMTDA} assumes that there are no domain labels for targets and a common model can be optimized by minimizing the discrepancy between the source and clusters of pseudo sub-targets. This technique is, however, unsuitable for detection due to the lack of instances annotations for object clustering \cite{BlendMTDA}. More recently, MT-MTDA \cite{mt-mtda} was proposed, which significantly outperforms \cite{BlendMTDA} by training one specialized teacher per target domain and iteratively distilling each teacher to a common model (Fig.\ref{fig:mtda_scenario}~(c)). While \cite{mt-mtda} can be extended to detection by simply having one teacher detector per target and then apply feature distillation, it is unsuitable for real-world application due to the high computational requirements of \cite{mt-mtda}, where one teacher detector per target domain is required. Furthermore, these methods are inflexible since they require access to previous target data when adapting for a new target domain. These limitations impede the deployment of these models in scenarios with limited storage or for privacy concerns on previous target domains. 

Motivated by above issues with current MTDA strategies (i.e., low level of performance or high complexity), this paper focuses on low-cost incremental MTDA strategies for object detection tasks.  Unsupervised incremental learning is leveraged with the goal of adapting a common model that can generalize well across multiple target domains without labels and without having to retrain on previously learned target data (see Fig. \ref{fig:mtda_scenario}~(d)). To be the best of our knowledge,~ \cite{wei2020_inc_mtda} is the only other method proposed to tackle incremental MTDA for the detection task. When adapting to a new target domain, it relies on knowledge distillation with a duplicated model trained on source data. This method distills knowledge from this duplicated to a common model, and constrains the output of the common model such that it does not deviate too much from the duplicated model. While this approach can alleviate knowledge corruption (i.e., catastrophic forgetting), its improvement in terms of detection accuracy remains limited, as it does not employ source-target feature alignment after adaptation, nor existing domain discriminators at different levels -- image and instances (i.e., bounding box level) -- of the detector. As shown in \cite{IJCNN_KD_UDA}, the performance of this method may also decline because the duplicate model only uses source data for distillation, as it does not have access to previous target data. Lastly, this approach is costly in terms of resources since it requires a duplicate model for adaptation, making it less practical in real-world scenarios.

\begin{figure}[htbp]
    \centering
    \includegraphics[width=0.8\textwidth]{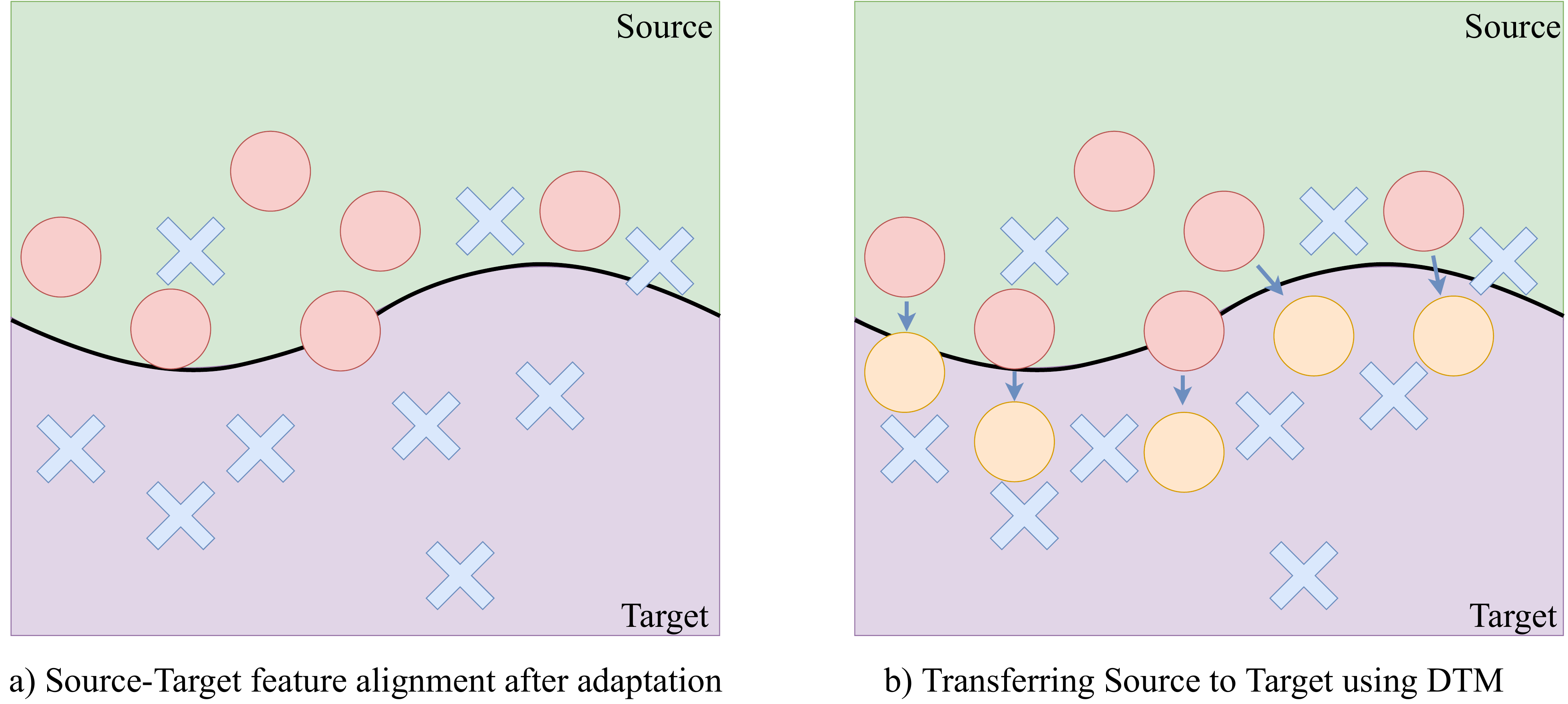}
    \caption{a) An illustration of the alignment of source (red circle) and target (blue cross) feature representations after domain adaptation. b) An illustration of the transfer of source feature (red circle) to pseudo-target feature representations (orange circle) that will be classified the same as target features (blue cross). Best viewed in color.}
    \label{fig:boundary}
    \vspace{-3mm}
\end{figure}

In order to address the aforementioned limitations, we propose an efficient MTDA with a Domain Transfer Module (MTDA-DTM). Our method allows training a common detector incrementally such that can achieve good generalization on multiple target domains. The proposed learning strategy allows domain adaptation to new target domains, without using data from previously-learned target domains, resulting in a more suitable technique for realistic scenarios. To prevent catastrophic forgetting our model integrates a Domain Transfer Module (DTM) to transfer source domain images to a joint image representation space shared across previous target domains.  In particular, once a model has been adapted to a new domain, the features representation of source and target images should be similar as the feature extractor was trained to maximize domain confusion. As shown in Fig. \ref{fig:boundary}, the proposed DTM leverages this feature alignment and transfers source images to the target domains. It is optimized such that the DTM output will be classified by domain discriminators as "target". When adapting to a new target domain, the DTM is used to generate pseudo images from previous target data distribution, thus alleviating catastrophic forgetting. Once the adaptation to a new target domain is achieved, the DTM training can be repeated to create a new DTM that includes the latest target. One of the advantages of our approach is that DTM can be trained without constraints, such as that transferred source images need to be realistic. This contrasts with adversarial \cite{explaining_harnessing_adversarial_ex} or GAN generated \cite{CycleGAN2017, GAN} image samples, as discussed in \ref{fig:mask_samples}. Compared to finding adversarial examples, where the distance between original and generated images must minimize a distance metric (i.e. L0, L1, L2, etc.), DTM provides more freedom to optimize a domain-invariant representation that's close to previously-learned targets. MTDA-DTM improves upon \cite{wei2020_inc_mtda} by providing pseudo-target data drawn from the joint representation obtained with DTM, thereby improving accuracy since the approach proposed in \cite{wei2020_inc_mtda} only uses source data to prevent catastrophic forgetting. Finally, our method allows for a considerable reduction in computational complexity when compared to \cite{wei2020_inc_mtda} since DTM only needs 2 convolution layers as opposed to using a duplicated object detector to prevent catastrophic forgetting. 

The main contributions of this paper are summarized as follows.
\begin{itemize}
    \item An cost-effective method, referred to as MTDA with Domain Transfer Module (MTDA-DTM), which can adapt to novel target domains incrementally, while avoiding the problem of catastrophic forgetting.
    \item In particular, our approach integrates the new DTM, which transfers source images to a joint image space shared across previous target domains. This module  allows for efficient training and adaptation compared to relevant state-of-the-art methods \cite{wei2020_inc_mtda}.
    \item The proposed MTDA-DTM is compared extensively against state-of-the-art MTDA techniques for object detection, consistently outperforming these methods across public MTDA detection benchmarks: PascalVOC-Clipart-Watercolor-Comic \cite{PascalVOC, ClipWaterComic} and Cityscape-FoggyCityscape-RainCityscape \cite{cityscape,FoggyCityscape, RainCityscape}. Furthermore, we compared the proposed MTDA-DTM in a real-world video-surveillance application, addressing the problem of multi-camera person detection using the Wildtrack \cite{Wildtrack} dataset. Results show that our method brings a substantial improvement of up to 5\%, in terms of accuracy, on the popular PascalVOC benchmark, and up to 2\% on a real-world application dataset, i.e., Wildtrack.
\end{itemize}

\vspace{-3mm}
\section{Related Work}

\subsection{Single-Target Domain Adaptation (STDA) for Object Detection:}
While the literature for unsupervised domain adaptation in classification \cite{GRL, DA_PR_GAN} or segmentation\cite{DA_PR_seg_transla, DA_Pr_Seg_scale_inv} is well explored, adaptation to the object detection task is not straightforward since domain adaptation for regression differs greatly from classification. This is mostly due to the difference between representation, and losses used by classification and regression. Several methods exist in the literature for this specific task~\cite{HTCN, FRCNN_DA, Instance-Invariant_DA_OD}. In particular, authors in \cite{FRCNN_DA} propose to find domain-invariant features in both image and instances level using domain discriminators to encourage domain confusion through adversarial training. Most techniques have improved upon \cite{FRCNN_DA} by focusing on aligning local features such as color and texture \cite{saito2019strong}. \cite{uda_od_cross_domain_aligment} finds local region of interest for detection and aligns them. Recently, \cite{multiadversarial_frcnn} applied domain adaptation on multiple layers of the feature extractor in addition to a weighted gradient reversal layer to handle hard confused domain samples. Using data augmentation, \cite{diversify_frcnn} proposes to apply single target domain adaptation by finding multi-domain invariant features of randomly augmented source and target domains. \textcolor{black}{\cite{Uncertainty-Aware_DA_OD} proposes to evaluate the sample alignment using an uncertainty metric in order to apply conditional domain adaptation with curriculum learning. Similarly, \cite{Self_guided_DA_Det} proposes to evaluate the "hardness" of samples in order to formulate an "easy-to-hard" adaptation, in addition to hierarchical feature alignment. Lastly, \cite{Instance-Invariant_DA_OD} improves upon state-of-the-art methods by extracting domain-invariant features by disentangling them from domain-specific features.}

More recently, some techniques have introduced CycleGAN \cite{CycleGAN2017} to map source data to target domains and vice versa in order to progressively adapt a detector step by step \cite{HTCN, Progressive_DA_FRCNN}. While this can significantly improve the results of STDA for object detection, it is not scalable for MTDA approaches, due to the number of generated (intermediate) datasets that scales linearly with the number of target domains (i.e. for the case of one source and one target domain, it would generate four dataset in total \cite{HTCN} ). The actual number of intermediate/generated and original datasets is $3n + 1$, with $n$ the number of target domains, thus making CycleGAN unrealistic for MTDA approaches. In addition, CycleGAN needs to access to previous target data for training, whereas our proposed MTDA-DTM does not require data from previous targets for training.

\vspace{-3mm}
\subsection{Multi-Target Domain Adaptation:}
Current approaches for MTDA mainly focus on the classification task. These approaches can essentially be categorized into two groups: 1) methods using domain labels \cite{MTDA_Theoric, mt-mtda} as additional information or 2) approaches assuming domain labels are unavailable \cite{BlendMTDA, mt-mtda, dada}. Nevertheless, as in the case of STDA techniques, MTDA approaches designed for classification are not suitable for object detection. AMEAN \cite{BlendMTDA}, one of the first techniques proposed for MTDA classification, assumes that no domain labels are available and tries to discover multiple hidden target domains using clustering on classification images. While AMEAN works for classification, it cannot be directly extended for object detection since, in this scenario, the object is among the background and there are not bounding box annotations, making the clustering of objects significantly difficult. 

With the state-of-the-art MT-MTDA\cite{mt-mtda} technique for classification, one teacher classification model is used for each target domain with logits distillation for a common student. \textcolor{black}{This technique can be employed with or without domain labels.} While \cite{mt-mtda} can be extended to detection by simply having one teacher detector per target domain, and applying feature distillation instead of logits. However, this MTDA approach would be prohibitively costly for detection, since object detection models are already large by themselves. Furthermore, this approach is not scalable in the case of multiple target domains. \textcolor{black}{Recently, another MTDA technique \cite{MTDA_Segmentation} has been also proposed for semantic segmentation using multiple teachers, where each teacher is responsible for a target domain, and distill to a single student. Although this method performs well on semantic segmentation tasks, it cannot be easily transferred to object detection as it uses specific features for the segmentation task, e.g. probability map for distillation, to improve performance. While, both these techniques achieve good generalization over multiple target domains, they lack flexibility to adapt the student model to new target domains, and require access to all target domains for learning.}

\vspace{-3mm}
\subsection{Incremental Learning for Object Detection:}
Incremental or continual learning refers to the ability of a model to learn new data while preserving previously-learned knowledge, without accessing all the previously-learned data. 
This information can range from new classes, new tasks, or even new domains, such as in our case. In the current literature, there are three families of incremental learning. The first family of approaches is based on regularization using either weight-based \cite{EWC, IL_SI} or, more recently, based on knowledge distillation \cite{IL_Lwf, Hou_2019_CVPR}, which have shown to achieve better performance. Additionally, both types of regularization can also be combined \cite{IL_PC,IL_WR_KD} to further improve the performance. The second type of approaches is based on memory replay where a fraction of data from previous tasks is retained. The manner in which the data is kept can range from directly storing raw data \cite{MetaER}, using generative networks such as GAN or VAE to learn previous data distribution \cite{IL_RH_Vae}. The third is based on architecture design \cite{IL_AR_1, IL_AR_2} where the architecture of the CNN models can be adapted for new information. This can range from adding weights or even duplicated CNNs to an existing architecture.

While these techniques can perform well on supervised classification, they are ill-suited for tasks such as unsupervised detection due to the missing label information that is needed to select weights/samples or storing unsuitable information in detection. For instance, DER \cite{darER}, employs logits and even labels of certain samples to distill information from previous tasks. This is not possible in the context of object detection, where logits only help to mitigate catastrophic on the classification module and there is nothing to prevent the bounding box proposal module (RPN) to degrade on previously-learned knowledge. 

Even though a few attempts have been presented in few-shot detection\cite{Inc_few_shot_sup}, current literature in unsupervised incremental is very limited. The work of \cite{Inc_OD_2017_ICCV} handles the problem of incremental learning in object detection by introducing a replica of the detector trained on the previous task. This detector serves as a teacher and preserves knowledge of the previous task when the current detector learns a new task using distillation. \cite{wei2020_inc_mtda} employs the approach in \cite{Inc_OD_2017_ICCV} in an unsupervised manner for MTDA by re-using features from the duplicated detector for domain confusion and a L2 distillation on outputs to avoid catastrophic forgetting. While \cite{wei2020_inc_mtda} achieves good performance for MTDA, we argue that it does not take full advantage of the current UDA techniques and both these techniques require substantial resources. In addition, it only uses the source domain for distillation which has been shown in\cite{IJCNN_KD_UDA} to have limited knowledge transfer and can reduce accuracy. Also, this distillation with source only guarantees consistency w.r.t source domain and does not guarantee that features will remain domain-invariant with previous target domains. In this work, we take advantage of the close nature of the source and target features thanks to UDA, along with domain discriminators to produce synthetic examples from the joint representation of previous target domains to avoid knowledge corruption, while only introducing a slight overhead.

\vspace{-3mm}
\subsection{Multi-Task Learning for Object Detection:}
Currently, most works on Multi-Task Learning (MTL) for object detection \cite{mtl_self_supervised_od, mtl_spotnet} seek to improve the performance of the main task (detection) instead of aiming at performing well across all tasks. In addition, these techniques assume either the datasets are labeled, or that they belong to the same domain, thus allowing the use of semi-supervised techniques. This contrasts with our learning strategy, which does not make any of these assumptions. Recently, \cite{multi_task_inc} proposes to tackle the problem of multi-domain detection. Their approach uses incremental learning by keeping samples from previous tasks to be used with distillation. 
This approach is not only computationally costly, but it also requires all task/target datasets to be labeled.

\vspace{-3mm}
\section{Proposed approach}

\vspace{-2mm}
\subsection{Preliminaries on Domain Adaptation for Object Detection:}

Let us define a source labeled dataset $S=\{(\XX,\yy)\}$, where $\XX_i \in \real^{H \times W \times 3}$ represents the \textit{i}-th image, and $H$ and $W$ its spatial dimension. Furthermore, $\yy_i$ denotes the corresponding ground truth vector, which contains 5 elements, $\yy_i = [x_i,y_i,w_i,h_i,c_i]$, with $x,y$ defining the top-left coordinates of the bounding box, $w,h$ its width and length, and $c$ denotes the corresponding class label ($c \in C$). In addition, we have access to an ensemble of multiple target domains $T=\{T_1, T_2,..., T_n\}$, where each target domain contains only unlabeled images, i.e., $T_i=\{\XX\}$. A feature extractor, $\phi$, is used to output a non-flattened feature map $\zz \in \real^{H_f \times W_f \times C_f}$, where $H_f$, $W_f$ and $C_f$ represent the feature representation dimension from an input image $\XX$, so that $\zz=\phi(\XX)$. Further, a function $r(\cdot)$ is used to extract instance-level features using the region proposal network (RPN) as in \cite{frcnn}, resulting into a vector $\pp \in \real^{V}$ where $V$ is an arbitrary size \cite{frcnn}. Last, a detector parameterized by $\Phi$ is used to provide the predicted vector $\hat{\yy}$ from training samples in $S$. Furthermore, since our problem focus on unsupervised targets, we consider that $\Phi(\XX)$ also used to provide $\zz$ and $\pp$ from training samples in $T_i$.


Currently, most state-of-the-art UDA techniques for object detection is based on \cite{FRCNN_DA}, where domain-invariant features are found by encouraging domain confusion on both image-level (global feature map produced by feature extractor) and instance-level (feature map of bounding boxes) with losses: \\
\vspace{-4mm}
\begin{equation}\label{eq:img_da_loss}
\mathcal{L}_{Img-DA}(\Phi, D_{img}, T_1) = \sum_{\XX \in S \cup T_1}\mathcal{L}_{C}(D_{img}(\zz), d)
\end{equation}
\begin{equation}\label{eq:inst_da_loss}
\mathcal{L}_{Inst-DA}(\Phi, D_{inst}, T_1) = \sum_{\XX \in S \cup T_1}\mathcal{L}_{C}(D_{inst}(\pp), d)
\end{equation}
\noindent Where $T_1$ indicate (first) target domain and $\zz$ and $\pp$ are obtained from $\XX$ using $\Phi$. $\mathcal{L}_C$ is a binary classification loss (e.g. focal loss \cite{FocalLoss} or standard cross-entropy)  and the domain label $d$ with $d=0$ indicates source domain and $d=1$ for target. $D_{img}$ and $D_{inst}$ are domain discriminators for the image-level and instance-level, respectively. We note that \cite{FRCNN_DA} also uses a consistency loss between image and instances. However, recent works \cite{HTCN, saito2019strong} have found that including this consistency term does not provide any improvement. Let us assume that a pre-trained detector is adapted to a target domain $T_1$ based on source data $S$, the overall domain adaptation loss for detection to update parameters of $\Phi$ and $D_{img}$ and $D_{inst}$ are characterized as:
\begin{equation}\label{eq:frcnn_da_loss}
\begin{aligned}
\mathcal{L}_{DA}(\Phi, D_{img}, D_{inst}, T_1) = \lambda (\mathcal{L}_{Img-DA}(\Phi, D_{img}, T_1) \\ + \mathcal{L}_{Inst-Da}(\Phi,  D_{inst}, T_1))  + \mathcal{L}_{Sup}(S)
\end{aligned}
\end{equation}
\noindent where $\mathcal{L}_{Sup}$ is the supervised loss on labeled source $S$, defined as $\mathcal{L}_{Sup} = \mathcal{L}_{reg} + \mathcal{L}_{cls}$, with $\mathcal{L}_{reg}$ the bounding box regression and $\mathcal{L}_{cls}$ the object classification losses. Further detail can be found in \cite{frcnn}. Parameter $\lambda$ controls the importance of domain adaptation losses.  Recent works like  \cite{saito2019strong} or \cite{HTCN} (used in our experiments) also have additional domain adaptation losses on different feature levels of the feature extractor. For the simplification of notation, we consider all domain discriminators (i.e. $D_{img}$, $D_{inst}$ or others in \cite{HTCN}) as $D_C$ and all domain adaptation losses as $\mathcal{L}_{OD-DA}$ since they only differ on whether its images or instances features. In addition, since bounding boxes and labels are only important for evaluation on source domain supervised training, we omit them in our notation:
\begin{equation}\label{eq:da_loss}
\mathcal{L}_{DA}(\Phi, D_C, T_1) = \lambda \mathcal{L}_{OD-DA}(\Phi, D_C, T_1) + \mathcal{L}_{Sup}(S)
\end{equation}

\subsection{Preliminaries on Incremental Learning:}

For incremental learning to a new target $T_2$, a basic approach involves applying Eq. \ref{eq:da_loss} for unsupervised domain adaptation of the detection model, on a new target data $T_2$:
\begin{equation}\label{eq:uft_da_loss}
\mathcal{L}_{UFT}(\Phi_{1}, D_{C_{1}}, T_2) = \lambda \mathcal{L}_{OD-DA}(\Phi_{1}, D_{C_1}, T_2) + \mathcal{L}_{Sup}(S)
\end{equation}
\noindent where $\Phi_{1}$ and $D_{C_1}$, are respectively, a detector and domain discriminators that are already adapted to previous target domains ($T_1$), and ($T_2$) is the new target to be learned. 

While this simple approach can work because the source domain can serve as an anchor for all the targets, it may still skew the detector toward the latest target domain. Indeed, using the source domain as an anchor only has a limited ability to prevent catastrophic forgetting since there are no explicit constraints on previous targets. To overcome this, we can retain the data of the previously-learned target domain, $T_1$:
\begin{equation}\label{eq:UInc_da}
\mathcal{L}_{UFT-Prev}(\Phi_1, D_{C_{1}}, T_1 \cup T_2) =  \lambda \mathcal{L}_{OD-DA}(\Phi_{1}, D_{C_1}, T_1 \cup T_2) + \mathcal{L}_{Sup}(S)
\end{equation}

However, it is prohibitively costly in many real-world detection applications to store and retrain on a mixture of data from all domains, especially with a growing number of target domains. In addition, incremental learning scenarios seek to learn new data without having access to previously-learned data.

\vspace{-3mm}
\subsection{MTDA with Domain Transfer Module:}

With our proposed approach -- MTDA-DTM, we perform unsupervised incremental learning of new target data using pseudo data of previous targets. This data is obtained via our DTM that is trained to trick the domain discriminators into classifying source-transferred images as target. Once the DTM is trained, the detector can be adapted incrementally to a new target domain. The rest of this section details the training of DTM and our approach for incremental domain adaptation.

\begin{figure*}[ht!]
    \centering
    \includegraphics[width=0.9\textwidth]{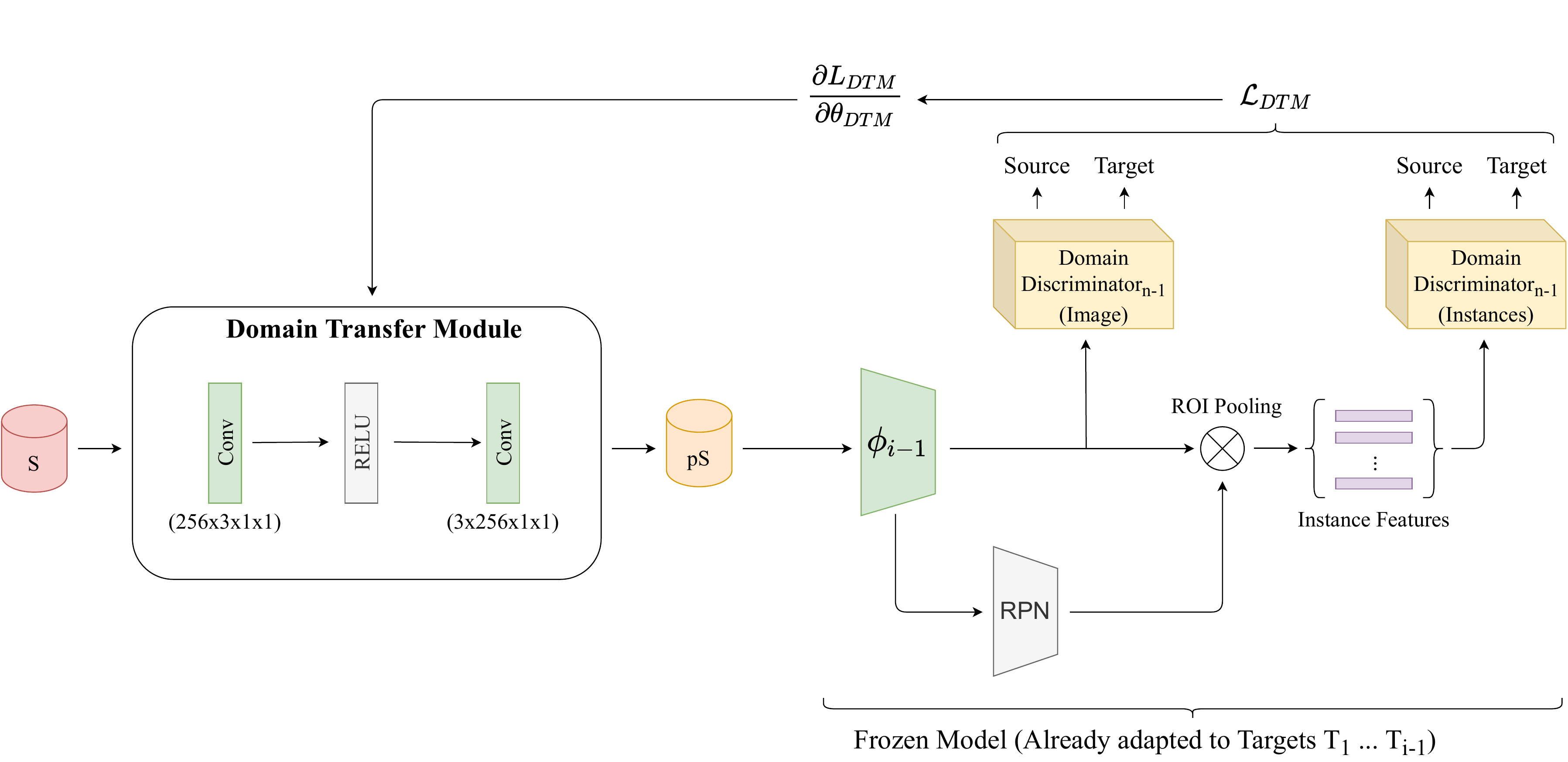}
    \captionsetup{justification=centering}
    \caption{Training the DTM prior to incremental DA of the detector to data of target $T_i$. The detector is already adapted to targets $T_1 ... T_{i-1}$. The pseudo-samples (pS) are generated by DTM. The dimension of the convolutional weights tensor of DTM are presented as (out channels $\times$ in channels $\times$ height $\times$  width)}
    \label{fig:DTM}
    \vspace{-3mm}
\end{figure*}

In order to access previously-learned target domains, we rely on the DTM, which transfers source domain samples to a joint representation of target domains. The transfer is performed using two convolution layers and a ReLU layer in between. As shown in \ref{fig:DTM}, the first convolution layer provides an expansion of the image channel-wise in order to project the image into a higher dimension space, and the second layer reduces the number of channels such that the output has the same number of channels as an image. This architecture was chosen since it is the simplest possible architecture, thus, allowing us to show the performance gain despite its simplicity. The optimization of DTM depends on how well pseudo-samples transferred by DTM from the source can force the domain discriminators to classify them as targets. We start by defining $g(\cdot)$ as the transformation learned by our DTM which is achieved  by minimizing the binary classification loss with domain label $d=1$ when samples are produced by DTM: 
\begin{equation}\label{eq:dtm_loss}
\mathcal{L}_{DTM} = \sum_{\XX \in S}\mathcal{L}_{C}(D_{C_{i}}(\Phi_i(g(\XX)), d = 1)
\end{equation}
\begin{figure}[htbp]
    \centering
    \includegraphics[width=0.75\textwidth]{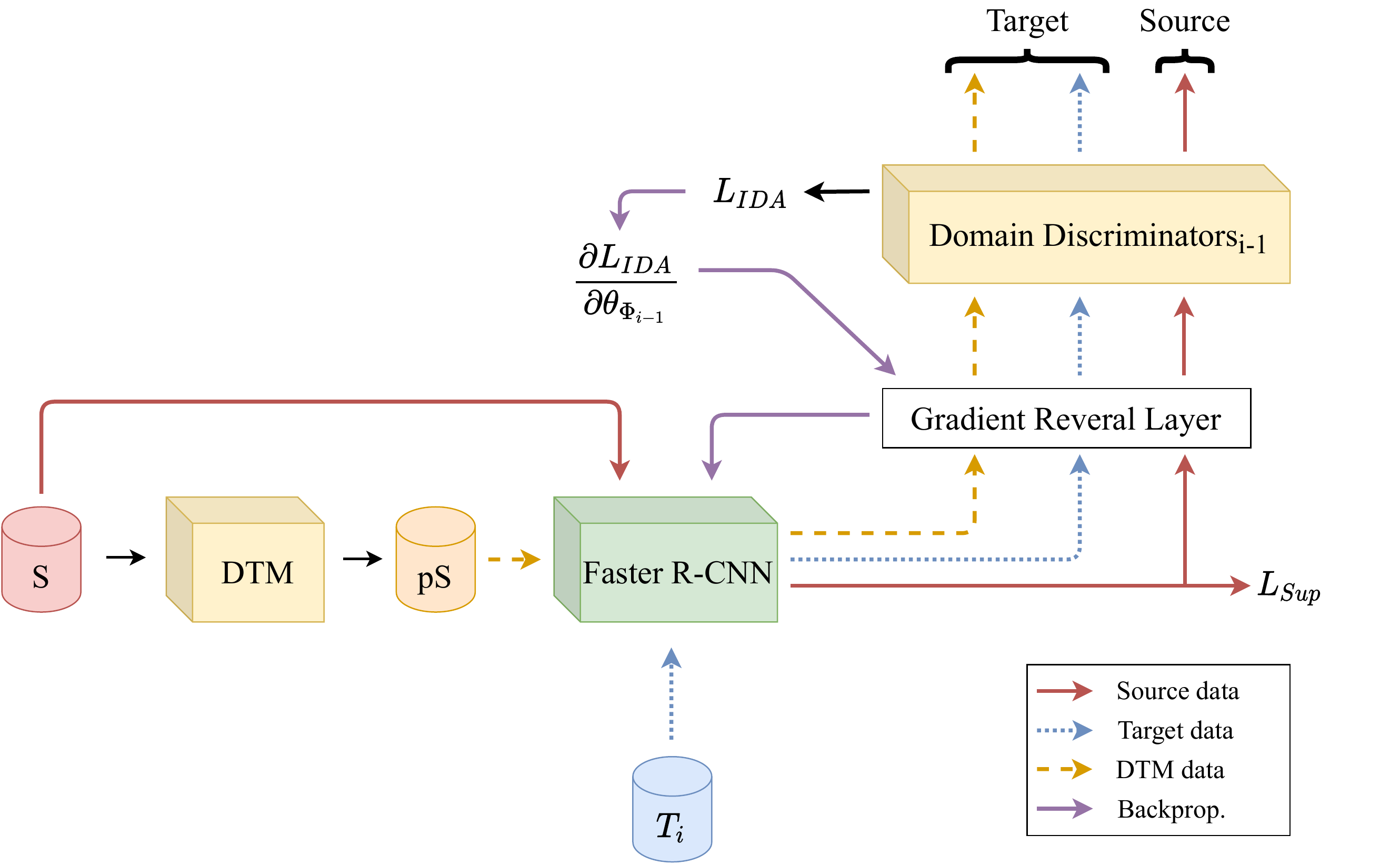}
    \captionsetup{justification=centering}
    \caption{Incremental adaption to target $T_i$ given a DTM already trained with data from targets $T_1 ... T_{i-1}$. $\theta_{\Phi}$ represents all the parameters (feature extractor, RPN, classifier) of our detector. }
    \label{fig:ida_dtm}
    \vspace{-3mm}
\end{figure}
\noindent where detector $\Phi_i$ and domain discriminator $D_{C_i}$ are already adapted to target $T_i$. Eq.\ref{eq:dtm_loss} is used to update the parameter of the DTM. Once the DTM is trained, the pseudo-samples transferred by DTM will be drawn from a joint representation of all targets, since $\Phi_i$ is adapted to $i$ previous target domain $\{T_1...T_i\}$ and its domain discriminators $D_C$ have been trained to distinguish between source vs $i$ prior target domains. Given that $D_{OD-DA}$ is just a simplification of other domain discriminators such as $D_{img}$, $D_{inst}$, etc., the pseudo-samples transferred by DTM will have both its domain transferred on both image and instance level providing both local and global translation. Also, pseudo-samples transferred by DTM are unbound by constraints such as the realism as in GAN, CycleGAN, etc., or affecting minimal changes (adversarial attack). Once the DTM is trained, previously-learned target domain in Eq.\ref{eq:UInc_da} can be replaced to use the learned $g()$ of DTM to prevent catastrophic forgetting: 
\begin{equation}\label{eq:ida}
\begin{aligned}
\mathcal{L}_{IDA}(\Phi_{i-1}, D_{C_{i-1}}, T_i) =  \lambda\sum_{\XX \in S \cup T_i}\mathcal{L}_{C}(D_{C_{i-1}}(\Phi_{i-1}(\XX)), d) \\ +  \alpha  \sum_{\XX\in S}\mathcal{L}_{C}(D_{C_{i-1}}(\Phi_{i-1}(g(\XX))), d=1) + \mathcal{L}_{sup}(S)
\end{aligned}
\end{equation}
\noindent where $\alpha$ represents the parameter to balance the importance of the loss on pseudo-samples from DTM. Fig. \ref{fig:ida_dtm} illustrates an incremental adaptation step  to a new target domain $T_i$ of our detection model with a DTM trained using Eq. \ref{eq:dtm_loss} with $\Phi_{i-1}$ . Compared to Eq.\ref{eq:UInc_da}, our optimization is more efficient since the DTM generates samples drawn from a joint representation of previously-learned targets, where there is less discrepancy between its samples compared to having different targets to optimize on as in Eq.\ref{eq:UInc_da}. Algorithm \ref{al:ida_dtm} presents the overall training strategy.

\begin{algorithm}[h]
\footnotesize
\SetAlgoLined
\caption{MTDA-DTM Training Strategy.}
\label{al:ida_dtm}
\SetKwInOut{Input}{Input}
\SetKwInOut{Output}{Output}
\SetKwInOut{Parameter}{parameter}
\Input{a source domain dataset $S$, a set of target dataset $T_1, ... T_n$ and a pretrained detection model $\Phi_0$}
\Output{a detection model $\Phi_n$, adapted to $n$ targets}

\For {$T_i \in (T_1, ... T_n)$} {
    \uIf{ $i = 1$}{
        STDA using Eq. \ref{eq:da_loss} \\
        Update domain discriminators $D_{C_0}$ and detection model $\Phi_{0}$ (feature extractor, RPN, classification and detection) \\
    }
    \uElse{
        Incr. DA using Eq. \ref{eq:ida} \\
        Update domain discriminators $D_{C_{i-1}}$ and detection model $\Phi_{i-1}$ (feature extractor, RPN, classification and detection) \\
    }
    Freeze the current detector model and domain discriminators \\
    Train a new DTM with $\mathcal{L}_{DTM}$ from Eq. \ref{eq:dtm_loss} using current domain discriminators $D_{C_i}$ and detection model $\Phi_{i}$
}

\end{algorithm}

\vspace{-3mm}
\section{Experimental Methodology}

\subsection{Datasets:}

\subsubsection{MTDA across datasets:} 

\paragraph{PascalVOC/Clipart/Watercolor/Comic:} This scenario regroups a set of well-known object detection benchmarks: PascalVOC (2007 + 2012) \cite{PascalVOC}, Clipart, Watercolor, and Comic \cite{ClipWaterComic}. In this scenario, PascalVOC is considered as the source, while and the three others as target domain datasets. PASCAL VOC 2007 is compiled of 2501 images for training, 2510 images for validation, and 4952 as test images, whereas PASCAL VOC 2012 contains 5717 images for training and 5823 images for evaluation. Clipart, Watercolor, and Comic have respectively 1000, 2000, and 2000 images, which are split to have $50\%$ for training and $50\%$ for the test set. For our scenario of MTDA with incremental learning, six common classes are selected among these datasets for training and evaluation. Examples from these datasets are shown in Fig.\ref{fig:data_samples}.

\vspace{-3mm}
\paragraph{Foggy/Rain/Cityscape:} This ensemble of datasets contains three different datasets: Cityscape \cite{cityscape} as the source, and FoggyCityscape \cite{FoggyCityscape} and RainCityscape \cite{RainCityscape} as target domains. In Cityscape, there are 3475 images of 8 categories, while FoggyCityscape provides 2500 images for training and 500 for test. As for RainCityscape, there are 9432 images for training and 1188 for test. Both FoggyCityscape and RainCityscape are synthetic datasets generated from Cityscape using \cite{FoggyCityscape} and \cite{RainCityscape}, with several image samples depicted in Fig.\ref{fig:data_samples}. In RainCityscape, we noticed a lack of samples for the class "train" in the evaluation subset. Thus, a set of $100$ randomly selected images with class "train" was extracted out of $500$ images in the training set and transferred to evaluation. We provide this list of images on our repository for the community. 

\subsubsection{MTDA across cameras:} 

\begin{figure}[htbp]
    \centering
    \includegraphics[width=0.8\textwidth]{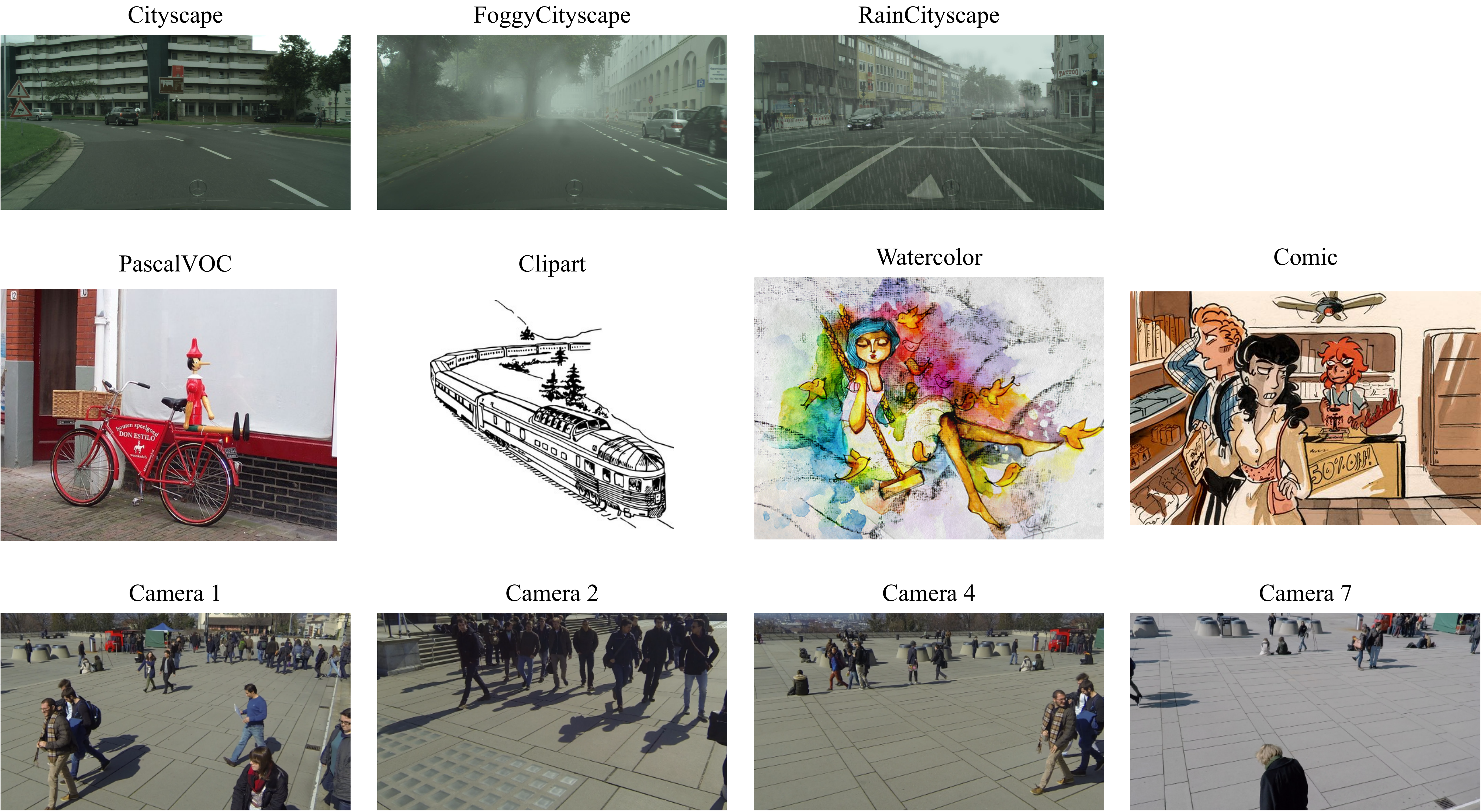}
    \captionsetup{justification=centering}
    \caption{Examples of images from different targets of each scenario. Best viewed in color.}
    \label{fig:data_samples}
    \vspace{-3mm}
\end{figure}

\paragraph{Wildtrack:} This scenario \cite{Wildtrack} is comprised of video data from seven different cameras made for supervised pedestrian detection and person re-identification. Each camera captured 400 frames at $1920\times1080$ resolution. This corresponds to a multi-camera DA scenario that's close to real-world applications. For our experiment on unsupervised IL, we will use Camera 1 (C1) as the source domain and all the other cameras (C2 $\xrightarrow{}$ C7) as target domains. Since the dataset is not provided with a standard split, a split of $2/3$ (train) and $1/3$ (test) was used for each camera to obtain a training set and an evaluation set, respectively. Fig.\ref{fig:data_samples} shows some image samples from Cameras (C1, C2, C4 and C7).

\vspace{-3mm}
\subsection{Implementation Details:}

For our experiments, for PascalVOC and Cityscape, we use the same settings as in \cite{HTCN} and other papers \cite{frcnn, FRCNN_DA}. The detection model is first trained with a learning rate of $0.001$ for 50k iterations, and then it is decreased by a factor of $10$ for the last 20k iterations. We use ResNet50 and VGG16 as a backbone CNN with weights pre-trained on ImageNet. For the domain adaptation technique, we use HTCN\cite{HTCN} for all our baselines. Since HTCN is currently the state-of-the-art for STDA in object detection and it can work without CycleGAN, our method is orthogonal to most techniques. In our incremental setting, starting from the second target domain adaptation, the learning rate will always be $0.0001$ and we run the algorithm for 70k iterations with the same learning rate scheduling. $\alpha$ is set as $0.1$ for the Cityscape related scenario as there is a smaller shift between Foggy/Rain/Cityscape, since, the only change in images is the synthetic weather. On the other hand, $\alpha$ is fixed to $1$ in the other settings, since the shift is larger between domains. This choice of $\alpha$ is also confirmed with a separate hold-out validation and its importance is evaluated in the Appendix B. Since our base STDA method is HTCN, we use the same value for $\lambda$ and other parameters such as UDA losses (attention losses, and instances). For Wildtrack scenario, we use a ResNet50 as the backbone, and the hyper-parameters differ with the PascalVOC only by the number of iterations: the detector is first trained for 4000 iterations with $0.001$ learning rate and then 1600 iterations with a learning rate reduced by a factor of $10$. Further details are provided in our Appendix A.

\vspace{-3mm}
\subsection{Baseline Models:}

Several baseline are compared to our method, they are divided into three categories: 1) lower bound, 2) incr. MTDA baselines, 3) ideal baselines and 4) fully-supervised baseline (upper bound). For the first category, there is \textit{Source Only} for the lower-bound, where a detection model is only trained on the source domain. For incr. MTDA baselines, the first baseline is \textit{UFT}, where the detector is fine-tuned through unsupervised domain adaptation to the next target. For ideal baselines, we consider the baselines shown in Figs.\ref{fig:mtda_scenario} a) and b). The baseline \textit{Only DA} represents a detector only adapted to one target domain using \cite{HTCN}, the baseline \textit{UFT Prev.}, similar to \textit{UFT} but with the addition of having access to previous target domains and lastly \textit{Mixed} baseline, where the detection model has access and is adapted to all the target domains at the same time. Lastly, when bounding box labels are available for \textcolor{black}{the training subsets} of target datasets, we provide supervised baselines (upper bound), such as: \textit{Only Sup.}, with supervised training of one detector per target, \textit{Supervised Mixed}, where the detector is trained in a supervised manner, directly on mixture of all target datasets, and \textit{Supervised FT}, where the detection model is fine-tuned in a supervised way from one target dataset to another. In terms of the state-of-the-art, we compare to Wei et al.\cite{wei2020_inc_mtda}, which we refer to as "Incr. MTDA KD" \footnote{No public code was provided, thus we use our own implementation for experiments.}. It uses distillation between a duplicated detector and the current detector to avoid catastrophic forgetting. Similar scenarios as in \cite{wei2020_inc_mtda} are used for comparison: Cityscape (\textbf{$S$})  $\xrightarrow{}$ FoggyCityscape (\textbf{$T_1$}) $\xrightarrow{}$ RainyCityscape (\textbf{$T_2$}) and PascalVOC (\textbf{$S$}) $\xrightarrow{}$ Clipart (\textbf{$T_1$}) $\xrightarrow{}$ Watercolor (\textbf{$T_2$}) while adding the target dataset Comic (\textbf{$T_3$}). For Wildtrack, we consider the following scenario: C1 (\textbf{$S$}) $\xrightarrow{}$ C2 (\textbf{$T_1$}) $\xrightarrow{}$ C3 (\textbf{$T_2$}) $\xrightarrow{}$ C4 (\textbf{$T_3$}) $\xrightarrow{}$ C5 (\textbf{$T_4$}) $\xrightarrow{}$ C6 (\textbf{$T_5$}) $\xrightarrow{}$ C7 (\textbf{$T_6$}). 

In all our experiments, we evaluate the accuracy of detection models according to the a mean average precision (mAP) with an IoU threshold = 0.5, where $N_C$ is the number of classes, and $AP_i$ is the average precision for class $i$:
\begin{equation}\label{eq:map}
mAP = \frac{1}{N_C}\sum_{i=0}^{N_C}AP_i
\end{equation}
To compare the resource requirements of detection models, we show the number of parameters of a detector for memory complexity, and the number of floating-point operations per second (FLOPS) for time complexity. All our experiments were performed using an Nvidia Tesla-P100 GPU.

\vspace{-3mm}
\section{Results and Discussion}

In this section, we first show the results of our method on MTDA benchmarks and then on the multi-camera problem. In a similar way to Incr. MTDA KD \cite{wei2020_inc_mtda}, results will be shown at each step and for each target on both scenarios of MTDA across datasets: Cityscape (\textbf{$S$})  $\xrightarrow{}$ FoggyCityscape (\textbf{$T_1$})  $\xrightarrow{}$ RainyCityscape (\textbf{$T_2$}),  and PascalVOC (\textbf{$S$})  $\xrightarrow{}$ Clipart (\textbf{$T_1$})  $\xrightarrow{}$ Watercolor (\textbf{$T_2$})  $\xrightarrow{}$ Comic (\textbf{$T_3$}). However, for the ablation study, we will only show the average accuracy of each target on each class. For the Wildtrack dataset, there is only one class ("pedestrian/person"), and we show the result at the end of incremental domain adaptations. Results at each incremental step of Wildtrack on Camera 2 are presented in an ablation study. In addition, a comparison with MT-MTDA \cite{mt-mtda} (adapted to detection) is also provide on the Cityscape scenario, since MT-MTDA \cite{mt-mtda} requires one teacher detection model per target, which makes it unusable for scenarios with multiple domains.

\vspace{-2mm}
\subsection{MTDA across datasets:}

\vspace{-2mm}
\paragraph{Cityscape (\textbf{$S$}) $\xrightarrow{}$ FoggyCityscape (\textbf{$T_1$}) $\xrightarrow{}$ RainCityscape (\textbf{$T_2$})}

\begin{table}[h!]
\centering
\caption{Average Precision of the proposed MTDA-DTM, baselines and state-of-the-art models for MTDA on the Cityscape (\textbf{$S$}) $\xrightarrow{}$ FoggyCityscape (\textbf{$T_1$}) $\xrightarrow{}$ RainCityscape (\textbf{$T_2$}) scenario.}
\label{tb:comp_mtda_Cityscape_FoggyCityscape_2_RainCityscape_eval_all}
\resizebox{0.9\textwidth}{!}{
\begin{tabular}{|l||r|r|r|r|r|r|r|r||r|}
	\hline
	\textbf{Backbone:} VGG16  & \multicolumn{9}{c|}{ \textbf{Accuracy} }       \\ 
	\textbf{Models}    &  \textbf{bus}           & \textbf{bicycle}       & \textbf{car}           &\textbf{m. cycle}       & \textbf{person}        & \textbf{rider}         & \textbf{train}         & \textbf{truck}         & \textbf{mAP}           \\ \hline \hline
	\multicolumn{10}{|l|}{ \textbf{Train: Cityscape (\textbf{$S$}) $\xrightarrow{}$ FoggyCityscape (\textbf{$T_1$}) $\xrightarrow{}$ RainCityscape (\textbf{$T_2$}) -- Test: FoggyCityscape (\textbf{$T_1$})}  } \\ \hline \hline
	Source Only                        & 24.4          & 28.4          & 27.0          & 16.4          & 24.5          & 31.0          & 9.2           & 13.3          & 21.8          \\ \hline \hline
	UFT                                 & \textbf{47.6} & 34.8          & 44.5 & 28.5          & 29.9 & \textbf{45.5}          & \textbf{34.6} & 27.8 & 36.6 \\ \hline
	Incr. MTDA KD \cite{wei2020_inc_mtda}                           & 45.6          & \textbf{35.3}         & \textbf{44.7}          & 31.9          & \textbf{31.5}          & 44.0          & 29.3          & 25.8         & 36.0          \\ \hline
	MTDA-DTM (ours)                               & 46.9 & \textbf{35.3} & \textbf{44.8} & \textbf{32.2} & \textbf{31.5} & \textbf{45.5} & \textbf{34.4}          & \textbf{29.1}          & \textbf{37.5}          \\ \hline \hline
	Only DA \cite{HTCN}                             & 48.0          & 34.4          & 47.1          & 29.4          & 33.0          & 48.5          & 37.4          & 29.3          & 38.4          \\ \hline
	UFT Prev.                           & 45.2          & 34.9          & 44.8          & 29.9          & 32.0          & 47.9          & 33.1          & 27.0          & 36.8          \\ \hline
	Mixed                              & 44.6          & 32.8          & 44.1          & 25.1          & 31.6          & 43.2          & 36.5          & 28.4          & 35.8          \\ \hline \hline
	Only Supervised                             & 50.6          & 35.5          & 50.1          & 35.5          & 33.3          & 46.2          & 42.8          & 34.9          & 41.1          \\ \hline
	Supervised FT                           & 40.2          & 33.7          & 44.0          & 29.3          & 30.3         & 41.4          & 19.8          & 22.2          & 32.6          \\ \hline
	Supervised Mixed                              & 42.9          & 35.3          & 47.7          & 30.5          & 31.9          & 43.9          & 15.8          & 25.4          & 34.2          \\ \hline \hline
	
	\multicolumn{10}{|l|}{ \textbf{Train:  Cityscape (\textbf{$S$}) $\xrightarrow{}$ FoggyCityscape (\textbf{$T_1$}) $\xrightarrow{}$ RainCityscape (\textbf{$T_2$}) -- Test: RainCityscape (\textbf{$T_2$})} }   \\ \hline \hline
	Source Only                        & 63.7          & 24.8          & 41.6          & 5.4  & 23.3          & 53.8          & 23.9          & 9.8           & 30.8          \\ \hline \hline
	UFT                                 & 83.4 & \textbf{33.8}          & 51.7          & 8.9           & 26.1          & 61.1          & \textbf{36.4}          & 24.0          & \textbf{40.7}          \\ \hline
	Incr. MTDA KD \cite{wei2020_inc_mtda}                & 78.2          & 33.1 & \textbf{52.4} & \textbf{10.5}           & 26.1 & \textbf{61.6}          & \textbf{36.4}          & 23.4          & 40.2          \\ \hline
	MTDA-DTM (ours)                               & \textbf{85.1} & \textbf{34.0}          & 52.1          & 4.2           & \textbf{26.7} & \textbf{61.5} & \textbf{36.5} & \textbf{26.2} & \textbf{40.8} \\ \hline \hline
	Only DA \cite{HTCN}                             & 83.1          & 34.7          & 48.5          & 8.2           & 26.2          & 61.9          & 36.1          & 27.1          & 40.7          \\ \hline
	UFT Prev.                           & 80.7          & 31.4          & 52.2          & 4.8           & 26.0          & 61.7          & 36.1          & 23.7          & 39.6          \\ \hline
	Mixed                              & 80.2          & 34.3          & 48.6          & 10.0          & 25.7          & 59.0          & 37.3          & 20.8          & 39.5          \\ \hline \hline
	Only Supervised                             & 42.4          & 22.7          & 51.4          & 3.2          & 23.8          & 55.5          & 81.8          & 9.3          & 36.2          \\ \hline
	Supervised FT                           & 63.8          & 31.9          & 51.5          & 27.4          & 26.9         & 58.7          & 81.8          & 27.2          & 46.2          \\ \hline
	Supervised Mixed                              & 74.2          & 30.2          & 51.9          & 10.3          & 27.0          & 61.3          & 81.6          & 11.5          & 43.5          \\ \hline \hline
	\multicolumn{10}{|l|}{\textbf{Train: Cityscape (\textbf{$S$}) $\xrightarrow{}$ FoggyCityscape (\textbf{$T_1$}) $\xrightarrow{}$ RainCityscape (\textbf{$T_2$}) - Test: all targets} }                                                 \\ \hline \hline
	Source Only                        & 44.0          & 26.6          & 34.3          & 10.9          & 23.9          & 42.4          & 16.5          & 11.5          & 26.3          \\ \hline \hline
	UFT                                 & 65.5 & 34.3          & 48.1          & 18.7          & 28.0          & \textbf{53.3}          & \textbf{35.5} & 25.9          & 38.6 \\ \hline
	Incr. MTDA KD \cite{wei2020_inc_mtda} & 61.9          & 34.2          & \textbf{48.5} & \textbf{21.2}          & 28.8          & 52.8          & 32.8          & 24.6          & 38.1          \\ \hline
	MTDA-DTM (ours)                               & \textbf{66.0} & \textbf{34.6} & \textbf{48.4}          & 18.2 & \textbf{29.3} & \textbf{53.5} & \textbf{35.4}          & \textbf{27.6} & \textbf{39.1} \\ \hline \hline
	Only DA \cite{HTCN}                             & 65.5          & 34.5         & 47.8          & 18.8          & 29.6          & 55.2          & 36.7          & 28.2          & 39.5          \\ \hline
	UFT Prev.                           & 62.9          & 33.1          & 48.5          & 17.3          & 29.0          & 54.8          & 34.6          & 25.3          & 38.2          \\ \hline
	Mixed                              & 62.4          & 33.5          & 46.3          & 17.5          & 28.6          & 51.1          & 36.9          & 24.6          & 37.6          \\ \hline \hline
	Only Supervised                             & 65.0          & 34.3          & 48.4          & 18.6          & 29.1          & 53.7          & 35.9          & 27.6          & 39.1          \\ \hline
	Supervised FT                           & 52.0          & 32.8          & 47.7          & 28.3          & 28.6         & 50.1          & 50.8          & 24.7          & 39.4          \\ \hline
	Supervised Mixed                              & 58.5          & 32.7          & 49.8          & 20.4          & 29.4          & 52.6          & 48.7          & 18.4          & 38.8          \\ \hline 
\end{tabular}
}
\vspace{-3mm}
\end{table}

 Table \ref{tb:comp_mtda_Cityscape_FoggyCityscape_2_RainCityscape_eval_all} shows that our method performs slightly better than other baselines. We also noticed that "UFT" performs well for a naive solution. This suggests that existing STDA techniques are well-suited for incremental learning because it learned domain-invariant features. However, we also observe that on "UFT" and method of Wei et al. \cite{wei2020_inc_mtda}. "Incr. MTDA KD", the performance on previous target FoggyCityscape is lower than ours. This shows that our DTM is capable of generating samples that are close to previously-learned target of FoggyCityscape, thus our method has better abilities to limit the effects of catastrophic forgetting, therefore, increasing the accuracy. While the "Only DA" baseline has a better performance accuracy (it uses one detection model per target) because there are more parameters overall to optimize since each detector can specialize to a target, and it also requires significantly more resources. As for \textit{Mixed} and \textit{UFT Prev}, their performance is lower than MTDA-DTM due to the difficulty of directly generalizing on multiple target domains simultaneously. This difficulty of generalization is because target domains are far from each other which makes common domain-invariant features more difficult to find compared to finding domain-invariant for one target at a time. Compared to supervised baselines, our method can achieve comparable or even better accuracy in several cases. Results suggest that it is easier to generalize to one target domain at a time. In addition, having multiple target domains may help improving performance since the results of "Only Sup." in RainCityscape is lower than other baselines that have multiple target domains. Compared to \textit{Supervised FT}, where the performance is skewed toward RainCityscape, and suffers performance degradation in FoggyCityscape, our MTDA-DTM achieves high performance on both target domains. In the next scenario, the difference between our methods and unsupervised baselines is even more pronounced as the number of target domains grows.

\vspace{-3mm}
\paragraph{PascalVOC (\textbf{$S$}) $\xrightarrow{}$ Clipart (\textbf{$T_1$}) $\xrightarrow{}$ Watercolor (\textbf{$T_2$})}

\begin{table}[t!]
\centering
\color{red}
\caption{Average precision of the proposed MTDA-DTM, baselines and state-of-the-art models for MTDA on the PascalVOC $\xrightarrow{}$ Clipart $\xrightarrow{}$ Watercolor scenario.}
\label{tb:comp_mtda_voc_clipart_2_watercolor_eval_clipart_watercolor}
\resizebox{0.9\textwidth}{!}{
\begin{tabular}{|l||r|r|r|r|r|r||r|}
\hline
\textbf{Backbone}: Resnet50 \ \ \ & \multicolumn{7}{c|}{\textbf{Accuracy}}       \\ 
\textbf{Models}             & \textbf{bicycle}                                                  & \textbf{bird} & \textbf{car}  & \textbf{cat}  & \textbf{dog}  & \textbf{person} & \textbf{mAP}  \\ \hline \hline
\multicolumn{8}{|l|}{\textbf{Train: PascalVOC (\textbf{$S$}) $\xrightarrow{}$ Clipart (\textbf{$T_1$}) $\xrightarrow{}$ Watercolor (\textbf{$T_2$}) - Test: Clipart (\textbf{$T_1$})}}   \\ \hline
Source Only  & 44.7                                                     & \textbf{28.2} & 16.8 & \textbf{10.8} & 12.6 & 46.0   & 26.5 \\ \hline \hline
UFT           & 30.5                                                     & 24.5 & 24.1 & 3.1  & 8.3  & 57.6   & 24.7 \\ \hline
Incr. MTDA KD \cite{wei2020_inc_mtda}     & 29.1                                                     & \textbf{28.1} & 26.8 & \textbf{10.6} & \textbf{20.3} & \textbf{64.8}   & 29.9 \\ \hline
MTDA-DTM (ours)         & \textbf{51.2}                                                     & 26.3 & \textbf{28.4} & 3.7  & 17.4 & \textbf{64.7}   & \textbf{31.9} \\ \hline \hline
Only DA \cite{HTCN}         & 19.3                                                     & 25.1 & 25.9 & 2.0  & 7.8  & 62.5   & 23.8 \\ \hline 
UFT Prev.    & 35.1                                                         & 27.5     & 26.1     & 2.2     & 16.3     & 63.3       & 28.4      \\ \hline
Mixed        & 43.2                                                         & 25.9     & 23.2      & 14.1      & 7.1      & 61.4       & 29.1      \\ \hline \hline

Only Supervised      & 42.7                                                    & 41.1  & 41.3 & 21.6    & 29.9 & 70.6   & 41.0 \\ \hline 
Supervised FT    & 36.4                                                    & 30.7 & 17.1  & 7.8 & 15.1 & 55.6  & 27.1     \\ \hline
Supervised Mixed        & 41.1                                                     & 39.0 & 44.6  & 28.4 & 20.8 & 70.4   & 40.7 \\\hline \hline

\multicolumn{8}{|l|}{\textbf{Train: PascalVOC (\textbf{$S$}) $\xrightarrow{}$ Clipart (\textbf{$T_1$}) $\xrightarrow{}$ Watercolor (\textbf{$T_2$}) - Test: Watercolor (\textbf{$T_2$})}}  \\ \hline \hline
Source Only    & 62.7                                                     & \textbf{45.6} & 42.5 & 30.0 & 29.4 & 59.0   & 44.9 \\ \hline \hline
UFT             & 64.5                                                     & 43.9 & \textbf{49.3} & \textbf{30.8} & 32.5 & 58.6   & 46.6 \\ \hline
Incr. MTDA KD \cite{wei2020_inc_mtda}       & 74.1                                                     & 42.5 & 48.8 & 29.7 & \textbf{33.4} & 58.7   & \textbf{47.8} \\ \hline
MTDA-DTM (ours)            & \textbf{74.5}                                                     & 44.9 & 47.9 & 25.0 & 30.9 & \textbf{62.7}   &  \textbf{47.6} \\ \hline \hline
Only DA \cite{HTCN}            & 50.8                                                     & 47.5 & 48.4 & 40.0 & 30.9 & 56.3   & 45.6 \\ \hline 
UFT Prev.                 & 63.6                                                         & 44.8     & 52.5     & 31.1     & 31.4     & 61.8       & 47.5     \\ \hline 
Mixed                     & 54.4                                                         & 47.6     & 49.2      & 33.6      & 30.2     & 62.0        &  46.2    \\ \hline \hline

Only Supervised      & 57.8                                                    & 49.4  & 42.6 & 40.6    & 36.5 & 68.3   & 49.2 \\ \hline 
Supervised FT    & 62.5                                                    & 47.6 & 31.3  & 36.6 & 39.3 & 66.1  & 47.2     \\ \hline
Supervised Mixed        & 65.5                                                     & 52.3 & 45.7  & 36.7 & 38.2 & 68.2   & 51.1 \\\hline \hline

\multicolumn{8}{|l|}{\textbf{Train: PascalVOC (\textbf{$S$}) $\xrightarrow{}$ Clipart (\textbf{$T_1$}) $\xrightarrow{}$ Watercolor (\textbf{$T_2$}) - Test: all targets}}  \\ \hline \hline
Source Only  & 53.7                                                     & 36.9  & 29.6 & 20.4  & 21.0    & 52.5   & 35.7 \\ \hline \hline 

UFT         & 47.5                                                     & 34.2  & 36.7  & 16.9 & 20.4  & 58.1   & 35.6 \\ \hline
Incr. MTDA KD \cite{wei2020_inc_mtda}     & 51.6                                                     & 35.3  & 37.8  & \textbf{20.1} & \textbf{26.8} & 61.7  & 38.9 \\ \hline
MTDA-DTM (ours)          & \textbf{62.8}                                                    & \textbf{35.6} & \textbf{38.1} & 14.3 & 24.1 & \textbf{63.7}   & \textbf{39.8} \\ \hline \hline
Only DA \cite{HTCN}      & 35.0                                                    & 36.3  & 37.1 & 21.0    & 19.3 & 59.4   & 34.7 \\ \hline 
UFT Prev.    & 49.3                                                    & 36.1 & 39.3  & 16.6 & 23.8 & 62.5  & 37.9      \\ \hline
Mixed        & 48.8                                                     & 36.7 & 36.2  & 23.8 & 18.6 & 61.7   & 37.6 \\\hline \hline

Only Supervised      & 50.2                                                    & 45.4  & 41.9 & 31.1    & 33.3 & 69.4   & 45.2 \\ \hline 
Supervised FT    & 49.5                                                    & 39.1 & 24.2  & 22.2 & 27.2 & 60.8  & 37.2      \\ \hline
Supervised Mixed        & 53.2                                                     & 45.7 & 45.1  & 32.5 & 32.5 & 29.5   & 69.3 \\\hline

\end{tabular}
}
\vspace{-3mm}
\end{table}

Table \ref{tb:comp_mtda_voc_clipart_2_watercolor_eval_clipart_watercolor} reports the accuracy of MTDA-DTM on both Clipart ($T_1$) and Watercolor (the new target domain, $T_2$). Results indicate that it still performs better than MTDA baselines and the state-of-the-art, especially on Clipart. Regarding the performance on Watercolor (the second target domain), our method achieves results comparable to Incr. MTDA KD \cite{wei2020_inc_mtda}. Table \ref{tb:comp_mtda_voc_clipart_2_watercolor_eval_clipart_watercolor} also show the benefits of having multiple target domains. By comparing "UFT", Incr. MTDA KD \cite{wei2020_inc_mtda}, MTDA-DTM, and "Only DA", results show that having multiple targets can lead to better performance than one detection model per target. Comparing with "UFT Prev." and "Mixed", results again confirm our analysis that generalizing directly on multiple targets can reduce performance. \textcolor{black}{As for the supervised case (upper bounds), we confirm that the baselines with access to all domains, "Supervised Mixed," and that trains on one dataset, "Only Supervised," achieve higher performance thanks to class labels. However, performance degrades once there is fine-tuning to a new domain without having access to previously-learned domains in "Supervised FT".} For the next incremental step, the Comic dataset is learned, and the effectiveness of our MTDA-DTM in the presence of multiple domains can be observed.

\vspace{-3mm}
\paragraph{PascalVOC (\textbf{$S$}) $\xrightarrow{}$ Clipart (\textbf{$T_1$}) $\xrightarrow{}$ Watercolor (\textbf{$T_2$}) $\xrightarrow{}$ Comic (\textbf{$T_3$})}

\begin{table}[t!]
\color{red}
\centering
\caption{Average Precision of the proposed MTDA-DTM, baselines and state-of-the-art models for MTDA on the PascalVOC $\xrightarrow{}$ Watercolor $\xrightarrow{}$ Comic scenario.}
\label{tb:comp_mtda_voc_clipart_watercolor_2_comic_eval_wcc}
\resizebox{0.9\textwidth}{!}{
\begin{tabular}{|l||r|r|r|r|r|r||r|}
\hline
\textbf{Backbone}: Resnet50 \ \ \ \ \ \ \ \ \ \ \ \ \ \ \ \ \ \ \ \ & \multicolumn{7}{c|}{\textbf{Accuracy} }     \\ 
\textbf{Models} & \textbf{bicycle}  & \textbf{bird} & \textbf{car}  & \textbf{cat}  & \textbf{dog}  & \textbf{person} & \textbf{mAP}  \\ \hline \hline
\multicolumn{8}{|l|}{\textbf{Train: PascalVOC (\textbf{$S$}) $\xrightarrow{}$ Clipart (\textbf{$T_1$}) $\xrightarrow{}$ Watercolor (\textbf{$T_2$}) $\xrightarrow{}$ Comic (\textbf{$T_3$}) - Test: Clipart (\textbf{$T_1$})}}    \\ \hline \hline             
Source Only  & 44.7                                                                         & 28.2 & 16.8 & \textbf{10.8} & 12.6 & 46.0   & 26.5 \\ \hline \hline
UFT           & 34.7                                                                         & 29.8 & 29.7 & 2.0  & 8.7  & 62.5   & 27.9 \\ \hline
Incr. MTDA KD \cite{wei2020_inc_mtda}     & 42.2                                                                         & 25.5 & 23.7 & 1.1  & 14.2 & 63.7   & 28.4 \\ \hline
MTDA-DTM (ours)          & \textbf{50.7}                                                                         & \textbf{34.0} & \textbf{32.1} & 5.2  & \textbf{16.3} & \textbf{64.9}   & \textbf{33.9} \\ \hline \hline
Only DA \cite{HTCN}       & 19.3                                                                         & 25.1 & 25.9 & 2.0  & 7.8  & 62.5   & 23.8 \\ \hline 
UFT Prev.    & 41.2                                                                             & 23.4     & 28.7      & 10.6      & 15.4      & 55.7        & 29.2      \\ \hline
Mixed        & 43.7                                                                         & 28.9 & 25.9 & 11.3 & 10.2 & 57.7   & 29.6 \\ \hline \hline

Only Supervised      & 42.7                                                    & 41.1  & 41.3 & 21.6    & 29.9 & 70.6   & 41.0 \\ \hline 
Supervised FT    & 50.4                                                    & 22.4 & 16.6  & 21.2 & 21.0 & 60.0  & 31.9     \\ \hline
Supervised Mixed        & 58.4                                                     & 40.6 & 44.4  & 23.3 & 31.1 & 69.9   & 44.6 \\\hline \hline

 \multicolumn{8}{|l|}{\textbf{Train: PascalVOC (\textbf{$S$}) $\xrightarrow{}$ Clipart (\textbf{$T_1$}) $\xrightarrow{}$ Watercolor (\textbf{$T_2$}) $\xrightarrow{}$ Comic (\textbf{$T_3$}) - Test: Watercolor (\textbf{$T_2$})}}    \\ \hline \hline

Source Only  & 62.7                                                                         & 45.6 & 42.5 & 30.0 & 29.4 & 59.0   & 44.9 \\ \hline \hline
UFT           & \textbf{72.3}                                                                         & 46.5 & 48.0 & \textbf{30.7} & 29.1 & \textbf{63.9}   & \textbf{48.4} \\ \hline
Incr. MTDA KD \cite{wei2020_inc_mtda}     & 68.3                                                                         & 43.5 & 49.1 & 24.4 & 28.2 & 62.3   & 46.0 \\ \hline
MTDA-DTM (ours)          & 66.4                                                                         & \textbf{48.0} & \textbf{49.5} & 29.9 & \textbf{30.8} & 62.2   & 47.8 \\ \hline \hline 
Only DA \cite{HTCN}       & 50.8                                                                         & 47.5 & 48.4 & 40.0 & 30.9 & 56.3   & 45.6 \\ \hline 
UFT Prev.    & 60.7                                                                             & 40.4     & 46.1      & 25.1      & 21.0   & 53.3        & 41.1      \\ \hline
Mixed        & 67.2                                                                         & 47.4 & 52.0 & 32.7 & 33.3 & 61.4   & 49.0 \\ \hline \hline 

Only Supervised      & 57.8                                                    & 49.4  & 42.6 & 40.6    & 36.5 & 68.3   & 49.2 \\ \hline 
Supervised FT    & 38.1                                                    & 40.6 & 24.5  & 32.2 & 15.7 & 63.9  & 35.8     \\ \hline
Supervised Mixed        & 68.1                                                     & 52.1 & 47.5  & 47.6 & 40.7 & 73.5   & 54.9 \\\hline \hline

\multicolumn{8}{|l|}{\textbf{Train: PascalVOC (\textbf{$S$}) $\xrightarrow{}$ Clipart (\textbf{$T_1$}) $\xrightarrow{}$ Watercolor (\textbf{$T_2$}) $\xrightarrow{}$ Comic (\textbf{$T_3$} - Test: Comic (\textbf{$T_3$})}}    \\ \hline \hline
Source Only     & 28.5                                                                         & 12.4 & 13.7 & 13.5 & 12.5 & 34.5   & 19.2 \\ \hline \hline
UFT              & 21.8                                                                         & 17.2 & 28.5 & 10.8 & 17.7 & 47.0     & 23.8 \\ \hline
Incr. MTDA KD \cite{wei2020_inc_mtda}        & 28.1                                                                         & 19.9 & 27.1 & 11.5 & 23.0   & 53.5   & 27.2 \\ \hline
MTDA-DTM (ours)             & \textbf{38.9}                                                                         & \textbf{22.5} & \textbf{32.3} & \textbf{15.3} & \textbf{30.9} & \textbf{56.8}   & \textbf{32.8} \\ \hline \hline
Only DA \cite{HTCN}          & 35.3                                                                         & 18.3 & 24.8 & 11.6 & 19.8 & 49.5   & 26.5 \\ \hline 
UFT Prev.       & 25.1                                                                             & 15.9      & 26.8      & 11.0      & 23.4      & 44.9        & 24.5             \\ \hline
Mixed           & 24.6                                                                         & 20.0   & 26.6 & 6.3  & 21.8 & 48.9   & 24.7        \\ \hline \hline 

Only Supervised      & 38.3                                                    & 22.1  & 37.6 & 37.4    & 42.2 & 70.1   & 41.3 \\ \hline 
Supervised FT    & 40.7                                                    & 22.1 & 38.2  & 40.5 & 39.0 & 63.8  & 36.2     \\ \hline
Supervised Mixed        & 48.5                                                     & 31.0 & 43.6  & 48.6 & 47.9 & 72.0   & 48.6 \\\hline \hline

\multicolumn{8}{|l|}{\textbf{Train: PascalVOC (\textbf{$S$}) $\xrightarrow{}$ Clipart (\textbf{$T_1$}) $\xrightarrow{}$ Watercolor (\textbf{$T_2$}) $\xrightarrow{}$ Comic (\textbf{$T_3$}) - Test: all targets}}    \\ \hline \hline
Source Only & 45.3                                                                         & 28.7 & 24.3 & \textbf{18.1} & 18.2 & 46.5   & 30.2 \\   \hline \hline
UFT           & 42.9                                                                         & 31.2 & 35.4 & 14.5 & 18.5 & 57.8   & 33.4 \\ \hline
Incr. MTDA KD \cite{wei2020_inc_mtda}    & 46.2                                                                         & 29.6 & 33.3 & 12.3 & 21.8 & 59.8   & 33.8 \\ \hline
MTDA-DTM (ours)        & \textbf{52.0}                                                                         & \textbf{34.9} & \textbf{38.0} & 16.8 & \textbf{26.0}   & \textbf{61.3}   & \textbf{38.1} \\ \hline \hline
Only DA \cite{HTCN}      & 35.1                                                                         & 30.3 & 33.0   & 17.9 & 19.5 & 56.1   & 32.0 \\ \hline
UFT Prev.    & 42.9                                                                             & 28.1     & 36.4     & 18.0     & 22.2     & 49.1       & 32.8      \\ \hline
Mixed     & 45.2                                                                         & 32.1 & 34.8 & 16.8 & 21.8 & 56.0     & 34.4 \\ \hline \hline

Only Supervised      & 46.3                                                    & 37.5  & 40.5 & 33.2    & 36.2 & 69.7   & 43.8 \\ \hline 
Supervised FT    & 43.1                                                    & 28.3 & 26.4  & 31.3 & 25.2 & 62.6 & 36.2     \\ \hline
Supervised Mixed        & 58.3                                                     & 41.2 & 45.2  & 39.8 & 40.0 & 71.8   & 49.4 \\ \hline

\end{tabular}
}
\vspace{-2mm}
\end{table}

Table \ref{tb:comp_mtda_voc_clipart_watercolor_2_comic_eval_wcc} shows the accuracy when our previous detection model is incrementally adapted to a third target domain (Comic, $T_3$). While results show that our model has slightly lower accuracy for Watercolor, it provides a significant improvement on Clipart, up to $6\%$, and Comic, up to $9\%$ mAP. Compared to "Only DA", results still confirm our previous analysis that multiple target domains can help improve performance. Results indicate that incremental approaches such as Incr. MTDA KD \cite{wei2020_inc_mtda} and MTDA-DTM, can achieve a high level of accuracy by adapting to one target domain at a time compared to simultaneously adaptation. This increase in accuracy is due to the learned joint representation of DTM which proves to be more suitable for optimization since the generated samples are from the joint representation compared to samples that are drawn directly from different target domains with different underlying distributions. \textcolor{black}{Finally, results with the supervised baselines further confirm our previous observations.} Overall, our results for general MTDA benchmarks across datasets show that MTDA-DTM outperforms all unsupervised baselines and can have comparable performance to some upper bound supervised baselines.

\subsection{MTDA across cameras:}

\paragraph{Wildtrack C1 (\textbf{$S$}) $\xrightarrow{}$ C2 (\textbf{$T_1$}) $\xrightarrow{}$ ... $\xrightarrow{}$ C7 (\textbf{$T_6$}).}

\begin{table}[h]
    \centering
    \caption{Average Precision for Incremental Multi-Target Domain Adaptation from Wildtrack Camera 1 (\textbf{$S$}.) $\xrightarrow{}$ Camera 2 (\textbf{$T_1$}) $\xrightarrow{}$ ...  $\xrightarrow{}$ Camera 7 (\textbf{$T_6$}). }
	\label{tb:comp_mtda_wildtrack}
	\resizebox{0.9\textwidth}{!}{
		\begin{tabular}{|l||r|r|r|r|r|r||r|}
			\hline
			\textbf{Backbone}: Resnet50 \ \ \ \ \ \ \ \ \ \     & \multicolumn{7}{c|}{ \textbf{Accuracy} } \\ 
			Models                             &  \ \ \ \textbf{C2} \ \ \            &  \ \ \ \textbf{C3}  \ \ \           & \ \ \ \textbf{C4}   \ \ \          & \textbf{C5} \ \ \            &  \ \ \ \textbf{C6}   \ \ \          &  \ \ \ \textbf{C7}  \ \ \           &  \ \ \ \textbf{Average}       \\ \hline \hline
			\multicolumn{8}{|l|}{\textbf{Train: C1 (\textbf{$S$}) $\xrightarrow{}$ C2 (\textbf{$T_1$}) $\xrightarrow{}$ C3 (\textbf{$T_2$}) $\xrightarrow{}$ C4 (\textbf{$T_3$}) $\xrightarrow{}$ C5 (\textbf{$T_4$}) $\xrightarrow{}$ C6 (\textbf{$T_5$}) $\xrightarrow{}$ C7 (\textbf{$T_6$}) - Test: all targets}}                                                                                                      \\ \hline \hline
			Source Only                        & 41.5              & 57.5              & 51.9              & 29.4              & 33.2              & 63.0              & 46.1              \\ \hline \hline
			UFT                                 & 47.4          & 59.4          & 53.0          & 53.0          & \textbf{41.0} & 67.1          & 53.5          \\ \hline
			Incr. MTDA KD \cite{wei2020_inc_mtda} & 47.2          & 58.4          & \textbf{55.5} & \textbf{57.4} & 39.0          & 67.8          & 54.2          \\ \hline
			MTDA-DTM (ours)                                & \textbf{50.0} & \textbf{65.9} & 54.3          & \textbf{57.2} & \textbf{40.8} & \textbf{68.2} & \textbf{56.1} \\ \hline \hline
			Only DA                            & 47.4          & 55.9          & 51.6          & 57.9          & 38.8          & 65.8          & 52.9          \\ \hline
			UFT Prev.                             & 48.6              & 59.2              & 44.8              & 48.8              & 38.4              & 63.2              & 50.5              \\ \hline
			Mixed                              & 49.8              & 64.0              & 53.0              & 58.4              & 41.7              & 68.5              & 55.9              \\ \hline \hline
			Only Supervised & 61.8          & 70.6          & 59.1          & 72.2          & 50.7          & 80.5          & 65.8          \\ \hline
			Supervised FT                            & 47.1          & 58.3          & 55.7          & 71.3          & 24.4          & 80.6          & 56.2          \\ \hline
			Supervised Mixed                         & 60.3          & 69.5          & 60.0          & 74.0          & 49.5          & 77.5          & 65.1          \\ \hline
		\end{tabular}
	}
	\vspace{-2mm}
\end{table}

\begin{figure}[htbp]
    \centering
    \includegraphics[width=0.7\textwidth]{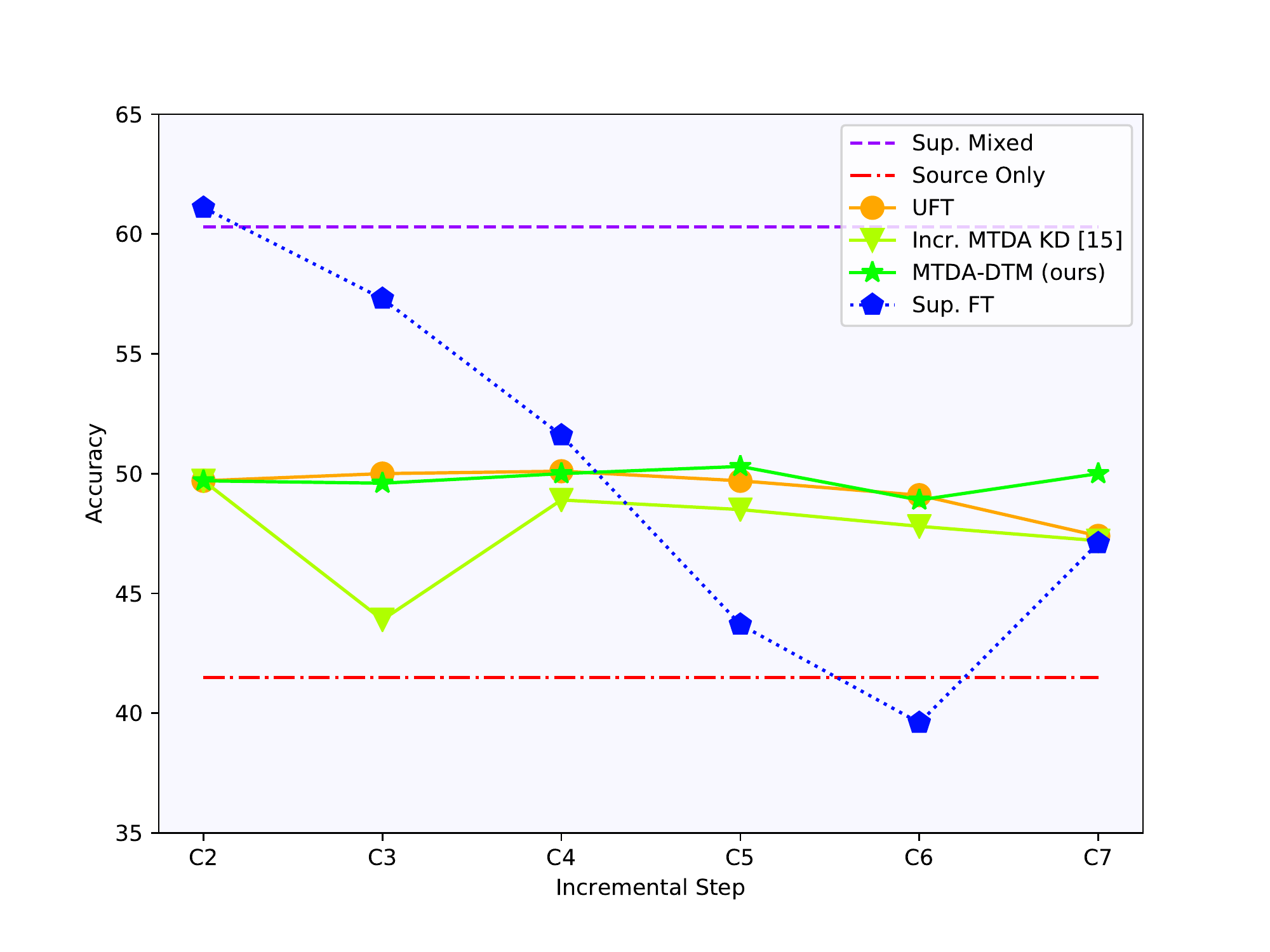}
    \caption{Evolution of detection accuracy on Wildtrack C2 at each incremental step. Best viewed in color. \protect \footnotemark}
    \label{fig:progressive_C2}
    \vspace{-3mm}
\end{figure}

Table \ref{tb:comp_mtda_wildtrack} confirms our results from previous tables and shows that our model outperforms comparable baselines and state-of-the-art, with performance gains of up to 2\% in terms of accuracy. Compared to several ideal unsupervised baselines, such as "Only DA" -{}-where one target is handled by one model-- or "Mixed" -{}-where all the target domains are mixed-{}-, our method also provides higher accuracy. This shows that using a common model for all the target domains and adapt to one target at a time can improve performance. Nevertheless, although our MTDA-DTM has lower accuracy than "Supervised Mixed', it is expected since the latter has access to labels. Our model has very similar accuracy to "Supervised FT", mainly due to catastrophic forgetting in "Sup.FT". The performance degradation can be seen in the evolution of accuracy of Camera 2 at each incremental step in the Figure \ref{fig:progressive_C2}. In addition, from the same figure, we again confirm that DTM provides better samples for optimization compared to using directly samples from previously-learned targets or using simultaneously samples of all the targets at once since MTDA-DTM mitigates catastrophic forgetting better than other methods.
\newline

 \vspace{-3mm}

\footnotetext{For ease of visualization, some non-incremental baselines such as Only DA, and Mixed were not included since they would remain the same as in Table \ref{tb:comp_mtda_wildtrack}}

From this figure,  we observe that most methods tend to have a decrease in accuracy between the first adaptation and the last. In addition, we notice a performance drop for Incr. MTDA KD \cite{wei2020_inc_mtda} at C3, which is due to its knowledge distillation although it managed to recover afterward thanks to learning from other domains. Comparing Incr. MTDA KD \cite{wei2020_inc_mtda} to "UFT" suggests that there's instability in the incremental learning process of Incr. MTDA KD \cite{wei2020_inc_mtda} since "UFT" does not suffer catastrophic forgetting even without any module to prevent. This further indicates that the distillation of only source samples can eschew the feature representation and cause instability to the model. This problem is better mitigated by our algorithm as we can see that it managed to maintain a stable accuracy over all the incremental steps.

\subsection{Ablation Studies:}

\paragraph{Comparison with MT-MTDA}

For this experiment, we adapted MT-MTDA \cite{mt-mtda}, one of the current state-of-the-art in MTDA for image classification, using a feature-based \cite{Overhaul} instead of logits-based distillation of the detection model. For MT-MTDA, we chose Resnet34 CNN backbone as teachers and Resnet18 as a common student to allow all the models to fit in memory. The same Resnet18 CNN backbone was used with our MTDA-DTM for a fair comparison. The Cityscape scenario was chosen for this experiment since MT-MTDA could not be run on either PascalVOC  or Wildtrack due to memory limitation (one model per target domain). Hyper-parameters used for MTDA-DTM are the same as the one used for Table \ref{tb:comp_mtda_Cityscape_FoggyCityscape_2_RainCityscape_eval_all}. As for MT-MTDA, hyper-parameters that give the best results were chosen using a validation set.

\begin{table}[h]
\centering
\caption{Comparison of the proposed and MT-MTDA \cite{mt-mtda} methods on the Cityscape scenario.}
\label{tb:comp_mt_mtda}
\resizebox{0.9\textwidth}{!}{
\begin{tabular}{|l||r|r|r|r|r|r|r|r||r|}
\hline
\textbf{Backbone}: Resnet18                                   & \multicolumn{9}{c|}{\textbf{Accuracy}}       \\ 
 \textbf{Models}                                             & \textbf{bus}                                              & \textbf{bicycle} & \textbf{car}   & \textbf{m.cycle} & \textbf{person} & \textbf{rider} & \textbf{train} & \textbf{truck} & \textbf{mAP}      \\ \hline \hline
\multicolumn{10}{|l|}{\textbf{Train: Cityscape (\textbf{$S$}) $\xrightarrow{}$ FoggyCityscape (\textbf{$T_1$}) $\xrightarrow{}$ RainCityscape (\textbf{$T_2$}) - Test: FoggyCityscape}}          \\
\hline \hline

MT-MTDA Res34 $\xrightarrow{}$ Res18 & 35.9                                             & 30.8    & 40.5  & \textbf{19.2}       & \textbf{28.4}   & 41.0    & 9.2   & 14.0    & 27.4   \\ \hline
MTDA-DTM (ours)                             & \textbf{42.8}                                             & \textbf{31.1}    & \textbf{43.6}  & 17.9       & 27.6   & \textbf{42.2}  & \textbf{10.7}  & \textbf{24.7}  & \textbf{30.1}   \\ \hline \hline
\multicolumn{10}{|l|}{\textbf{Train: Cityscape (\textbf{$S$}) $\xrightarrow{}$ FoggyCityscape (\textbf{$T_1$}) $\xrightarrow{}$ RainCityscape (\textbf{$T_2$}) - Test: RainCityscape}} \\ \hline \hline
MT-MTDA Res34 $\xrightarrow{}$ Res18             & 67.8                                             & \textbf{31.6}    & 48.4  & \textbf{4.9}        & 23.6   & 57.1  & 26.4  & 3.3   & 32.9  \\ \hline
MTDA-DTM (ours)                             & \textbf{72.8}                                             & 31.1    & \textbf{50.2}  & 2.7        & \textbf{24.0}     & \textbf{58.2}  & \textbf{31.4}  & \textbf{15.8}  & \textbf{35.8}   \\ \hline \hline
\multicolumn{10}{|l|}{\textbf{Train: Cityscape (\textbf{$S$}) $\xrightarrow{}$ FoggyCityscape (\textbf{$T_1$}) $\xrightarrow{}$ RainCityscape (\textbf{$T_2$}) - Test: all targets}}         \\ \hline \hline
MT-MTDA Res34 $\xrightarrow{}$ Res18             & 51.8                                            & \textbf{31.2}    & 44.4 & \textbf{12.0}      & \textbf{26.0}     & 49.0 & 17.8  & 8.6  & 30.1 \\ \hline
MTDA-DTM (ours)                               & \textbf{57.8}                                             & \textbf{31.1}    & \textbf{46.9}  & 10.3       & \textbf{25.8}   & \textbf{50.2}  & \textbf{21.0} & \textbf{20.2} & \textbf{32.9}   \\ \hline
\end{tabular}
}
\vspace{-3mm}
\end{table}

From Table \ref{tb:comp_mt_mtda}, we observe that our model outperforms MT-MTDA \cite{mt-mtda} on all datasets of the Cityscape scenario. Results show that an incremental approach, where target domains are adapted one at a time, can be beneficial compared to an approach trying to directly generalize on several domains even, when it uses compressed knowledge. \textcolor{black}{To further confirm these results, we adapt our technique to image classification, and evaluate on PACS dataset.} 
\textcolor{black}{Table \ref{tb:pacs_comp_mtda} shows that MTDA-DTM can outperform  MT-MTDA in a classification setting by adapting to one model at a time, instead of simultaneously adapting to multiple domains.}

\begin{table}[htbp]
\color{black}
\centering
\caption{Comparison of the proposed and \cite{MTDA_Theoric} methods on the PACS dataset for a classification setting.}
\label{tb:pacs_comp_mtda}
\resizebox{0.7\columnwidth}{!}{
\begin{tabular}{|l||rrrr|}
	\hline
	\textbf{Backbone:} LeNet & \multicolumn{4}{c|}{\textbf{Accuracy}}\\ \hline \hline
	
		  \multicolumn{5}{|c|}{\textbf{Train}: \textbf{P} (S) $\xrightarrow{}$ \textbf{Ap} ($T_1$) $\xrightarrow{}$ \textbf{Cr} ($T_2$) $\xrightarrow{}$ \textbf{S} ($T_3$) - \textbf{Test}: \textbf{all targets}}                                                                                                                                                                                                                        \\ \hline 
	
	\textbf{Models}                      & \multicolumn{1}{r|}{\textbf{Ap}}   & \multicolumn{1}{r|}{\textbf{Cr}}   & \multicolumn{1}{r||}{\textbf{S}}    & \multicolumn{1}{r|}{\textbf{Average}} \\ \hline \hline

	MT-MTDA Res50 $\xrightarrow{}$ LeNet & \multicolumn{1}{r|}{24.6}          & \multicolumn{1}{r|}{32.2}          & \multicolumn{1}{r||}{33.8}          & \multicolumn{1}{|r|}{30.2}             \\ \hline
	MTDA-DTM (Ours)                      & \multicolumn{1}{r|}{\textbf{52.0}} & \multicolumn{1}{r|}{\textbf{37.0}} & \multicolumn{1}{r||}{\textbf{39.6}} & \multicolumn{1}{r|}{\textbf{43.0}}     \\ \hline \hline
	
		 \multicolumn{5}{|c|}{\textbf{Train}: \textbf{Ap} (S) $\xrightarrow{}$ \textbf{Cr} ($T_1$) $\xrightarrow{}$ \textbf{S} ($T_2$) $\xrightarrow{}$ \textbf{P} ($T_3$) - \textbf{Test}: \textbf{all targets}}   \\ \hline 
	
	\textbf{Models}                                     & \multicolumn{1}{r|}{\textbf{Cr}}   & \multicolumn{1}{r|}{\textbf{S}}    & \multicolumn{1}{r||}{\textbf{P}}    & \textbf{Average}                       \\ \hline \hline

	MT-MTDA Res50 $\xrightarrow{}$ LeNet & \multicolumn{1}{r|}{46.6}          & \multicolumn{1}{r|}{\textbf{57.5}} & \multicolumn{1}{r||}{35.6}          & 46.6                                   \\ \hline
	MTDA-DTM (Ours)                      & \multicolumn{1}{r|}{\textbf{61.2}} & \multicolumn{1}{r|}{47.0}          & \multicolumn{1}{r||}{\textbf{83.0}} & \textbf{63.7}                          \\ \hline
		
\end{tabular}
}
\vspace{-4mm}
\end{table}

\paragraph{Open Domain Adaptation}

In this experiment, our model is evaluated in an open domain adaptation scenario, i.e. our model is evaluated on a target domain that was not employed for training/domain adaptation. In this open domain scenario, we assume that the "unseen" Comic would have the same number of classes and the same labels as our existing domains. Our model and baselines are trained on Pascal (\textbf{$S$}) $\xrightarrow{}$ Clipart (\textbf{$T_1$}) $\xrightarrow{}$ Watercolor (\textbf{$T_2$}) data and evaluated directly on Comic. In this scenario, "UDA Comic Only" represents a model trained with HTCN \cite{HTCN} from PascalVOC to only Comic without any intermediate targets.

\begin{table}[h!]
\centering
\caption{Average Precision for Open Domain Adaptation by evaluating models of \textbf{Pascal  (\textbf{$S$}) $\xrightarrow{}$ Clipart (\textbf{$T_1$}) $\xrightarrow{}$ Watercolor (\textbf{$T_2$})} on \textbf{Comic ($T_3$)}.}
\label{tb:comp_mtda_open}
\resizebox{0.8\textwidth}{!}{
\begin{tabular}{|l||r|r|r|r|r|r||r|}
\hline
\textbf{Backbone}: Resnet50  & \multicolumn{7}{c|}{\textbf{Accuracy}} \\ 
\textbf{Models}             & \textbf{bicycle}     & \textbf{bird}     & \textbf{car}      & \textbf{cat}      & \textbf{dog}      & \textbf{person}    & \textbf{mAP}     \\ \hline \hline
\multicolumn{8}{|l|}{\textbf{Train: PascalVOC (\textbf{$S$}) $\xrightarrow{}$ Clipart (\textbf{$T_1$}) $\xrightarrow{}$ Watercolor (\textbf{$T_2$}) - Test: Comic (\textbf{$T_3$})}}                                                    \\ \hline \hline
UFT           & 15.6        & 11.6     & 23.6     & 10.1     & 17.6     & 40.0      & 19.7    \\ \hline
Incr. MTDA KD\cite{wei2020_inc_mtda}     & 22.7        & 16.1     & 27.1     & 12.1     & 22.3     & 47.8      & 24.7    \\ \hline
MTDA-DTM (ours)         & \textbf{33.7}        & \textbf{20.5}     & \textbf{29.6}     & \textbf{14.6}     & \textbf{26.9}     & \textbf{55.0}      & \textbf{30.0}    \\ \hline
UFT Prev.     & 28.6        & 19.5     & 28.5     & 9.4      & 19.9     & 47.0      & 25.5    \\ \hline \hline
UDA Comic Only \cite{HTCN}       & 35.3        & 18.3     & 24.8     & 11.6     & 19.8     & 49.5      & 26.5    \\ \hline 
\end{tabular}
}
\vspace{-3mm}
\end{table}

Table \ref{tb:comp_mtda_open} shows that our model from Pascal (\textbf{$S$}) $\xrightarrow{}$ Clipart (\textbf{$T_1$}) $\xrightarrow{}$ Watercolor (\textbf{$T_2$}) outperforms all the other baselines on the unseen Comic target domain, including the baseline, "UDA Comic Only" that's only domain adapted to "Comic". Results shows that training with other diverse target domains data help improve the robustness of our features. Overall, results show that MTDA-DTM can provide robustness on an unseen target by providing pseudo-samples designed to trick the domain discriminator. 

\vspace{-3mm}
\paragraph{Impact of DTM}

In this case, the impact of using a trained DTM is compared to a non-trained DTM (i.e initial weights) on the scenario PascalVOC with three target domains. The goal of this experiment is to show the effectiveness of our adversarial training.

\begin{figure}[htbp]
    \centering
    \includegraphics[width=0.6\textwidth]{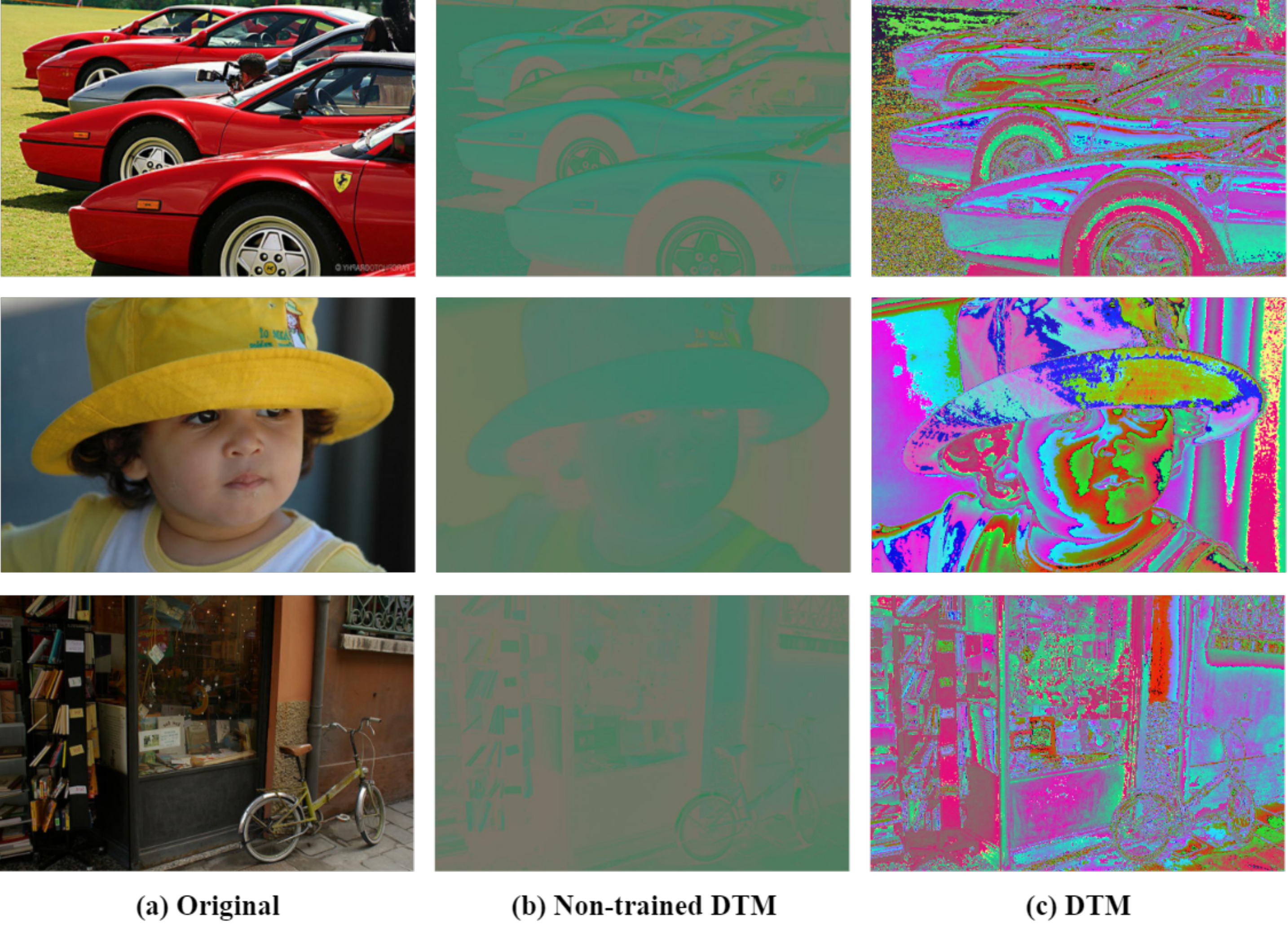}
    \caption{Examples of samples of DTM projected to image space (c) compared to original images (a) and a non-trained DTM (b) . Best viewed in color.}
    \label{fig:mask_samples}
\vspace{-5mm}
\end{figure}

\begin{figure}[htbp]
    \centering
    \includegraphics[width=0.6\textwidth]{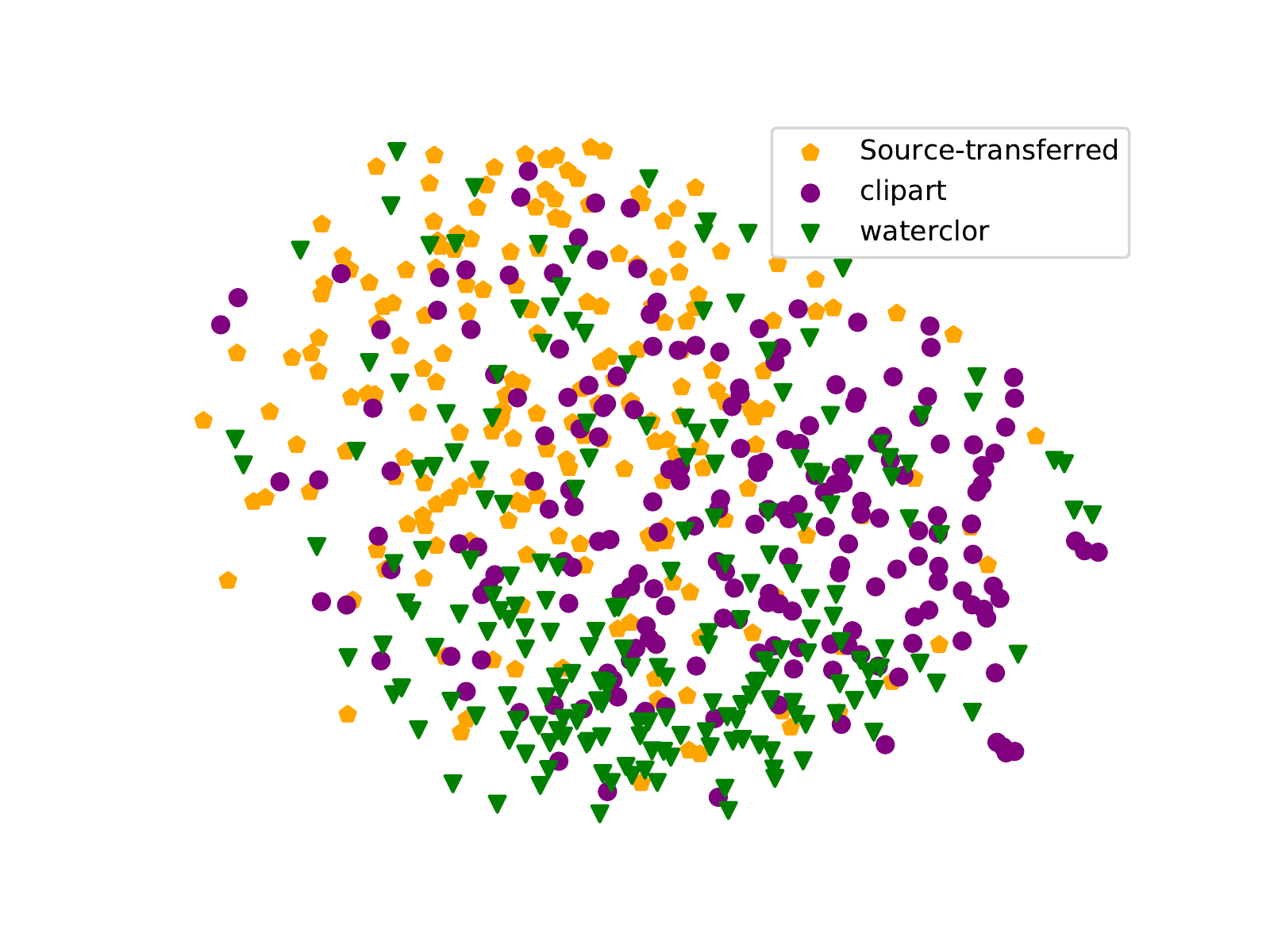}
    \caption{UMAP feature visualization of domain-transferred source domain (orange), all target domains: Clipart (purple) and Watercolor (green) from DTM trained for scenario PascalVOC (\textbf{$S$})  $\xrightarrow{}$ Clipart (\textbf{$T_1$})  $\xrightarrow{}$ Watercolor (\textbf{$T_2$})  $\xrightarrow{}$ Comic (\textbf{$T_3$}) . This figure shows the desired outcome of our DTM. Best viewed in color.}
    \label{fig:umap_mask_vs_source_vs_targets}
    \vspace{-2mm}
\end{figure}

\begin{table}[h]
\centering
\caption{Comparison of trained DTM vs not-trained DTM on Clipart, Watercolor, Comic.}
\label{tb:comp_dtm_vs_dtm_no_trained}
\resizebox{0.9\columnwidth}{!}{
\begin{tabular}{|l||r|r|r|r|r|r||r|}
\hline
 \textbf{Backbone}: Resnet50 \ \ \ \ \ \ \ \ \ \ \ \         & \multicolumn{7}{c|}{\textbf{Accuracy}} \\ 
\textbf{Models}                    & \textbf{bicycle}    & \textbf{bird}           & \textbf{car}      & \textbf{cat}            & \textbf{dog}      & \textbf{person}        & \textbf{mAP}           \\ \hline \hline                     
\multicolumn{8}{|l|}{\textbf{Train: PascalVOC  (\textbf{$S$}) $\xrightarrow{}$ Clipart (\textbf{$T_1$}) $\xrightarrow{}$ Watercolor (\textbf{$T_2$}) $\xrightarrow{}$ Comic (\textbf{$T_3$}) - Test: all targets}}                                                    \\ \hline \hline
UFT           & 15.6        & 11.6     & 23.6     & 10.1     & 17.6     & 40.0      & 19.7    \\ \hline
Incr. MTDA KD\cite{wei2020_inc_mtda}     & 22.7        & 16.1     & 27.1     & 12.1     & 22.3     & 47.8      & 24.7    \\ \hline \hline
MTDA-DTM Not Trained  & 48.3       & 30.5    & 36.9     & 14.4    & 24.9     & 56.2   & 35.2   \\ \hline
MTDA-DTM Trained  & \textbf{51.5}      & \textbf{34.8}          & \textbf{38.1}    & \textbf{16.8}          & \textbf{26.0}    & \textbf{61.3}         & \textbf{38.1}         \\ \hline
\end{tabular}
}
\vspace{-3mm}
\end{table}

In Fig. \ref{fig:mask_samples}, the pseudo-samples generated by DTM are projected back to image space in our Pascal (\textbf{$S$})  $\xrightarrow{}$ Clipart (\textbf{$T_1$})  $\xrightarrow{}$ Watercolor (\textbf{$T_2$})  $\xrightarrow{}$ Comic (\textbf{$T_3$})  scenario. Based on this figure, DTM seems to focus on light intensity of the images, which can be the underlying common features between target domains. Results of Table \ref{tb:comp_dtm_vs_dtm_no_trained} show that having an optimized DTM by using domain discriminators can achieve a significant increase in accuracy. In addition, "Not Trained" still has better accuracy than the other baselines. This is can be explained by the robustness of our detection model to open domain adaptation, as seen previously in Table \ref{tb:comp_mtda_open}. Fig. \ref{fig:umap_mask_vs_source_vs_targets} shows an UMAP \cite{umap} visualization of features from all target domains and the pseudo-samples using the feature extractor and DTM from PascalVOC (\textbf{$S$})  $\xrightarrow{}$ Clipart (\textbf{$T_1$}) $\xrightarrow{}$ Watercolor (\textbf{$T_2$}) . In this Figure, the features of samples from DTM, not only overlap with both previous target domains, which is our desired outcome, but they are also found outside of the distribution of previous targets, which can help improve accuracy. We employed UMAP \cite{umap} instead of TSNE \cite{TSNE} since it is better at preserving the global structure and it is also faster to compute.

\vspace{-3mm}
\paragraph{Order of incremental learning}

For this experiment, we evaluate the impact of the order of our incremental learning among the three target domains of Clipart (\textbf{Cl}), Watercolor (\textbf{W}), and Comic (\textbf{Co}) for the PascalVOC scenario and several pertinent target learning orders for Wildtrack scenario.

\begin{table}[h]
\centering
\caption{Overall accuracy per class with MTDA-DTM for different target adaptation orders on the PascalVOC scenario.}
\label{tb:comp_mtda_wcc_different_order}
\resizebox{0.8\columnwidth}{!}{
\begin{tabular}{|l||r|r|r|r|r|r||r|}
\hline
\textbf{Backbone}: Resnet50                                        & \multicolumn{7}{c|}{\textbf{Accuracy}}      \\ 
 \textbf{Order}  & \textbf{bicycle}                                                                        & \textbf{bird}  & \textbf{car}  & \textbf{cat}  & \textbf{dog}  & \textbf{person} & \textbf{mAP}  \\ \hline \hline
\multicolumn{8}{|l|}{\textbf{Train: Different Target Orders - Test: all targets}}       \\ \hline \hline
P $\xrightarrow{}$ Cl $\xrightarrow{}$ W $\xrightarrow{}$ Co & 51.5                                                                   & 34.8  & 38.1 & 16.8 & 26.0 & 61.3   & 38.1 \\ \hline
P $\xrightarrow{}$ Cl $\xrightarrow{}$ Co $\xrightarrow{}$ W & 53.3                                                                              & 30.9      & 37.4     & 16.4     & 25.1     & 60.3       & 37.2     \\ \hline
P $\xrightarrow{}$ Co $\xrightarrow{}$ Cl $\xrightarrow{}$ W & 54.7                                                                          & 30.1 & 37.7 & 20.2 & 24.5 & 57.2   & 37.4 \\ \hline
P $\xrightarrow{}$ Co $\xrightarrow{}$ W $\xrightarrow{}$ Cl & 47.8                                                                         & 29.0  & 40.0 & 15.5 & 23.4 & 58.8   & 35.7 \\ \hline
P $\xrightarrow{}$ W $\xrightarrow{}$ Cl $\xrightarrow{}$ Co & 53.7                                                                          & 29.0  & 31.1 & 17.4 & 18.0 & 55.7   & 34.2 \\ \hline
P $\xrightarrow{}$ W $\xrightarrow{}$ Co $\xrightarrow{}$ Cl & 49.8                                                                              & 31.3      & 33.1     & 18.4     & 20.7     & 56.4       & 34.9     \\ \hline
\end{tabular}
}
\vspace{-3mm}
\end{table}					

Table \ref{tb:comp_mtda_wcc_different_order} shows that there can be a significant change in accuracy depending on which target domain we start with. These results suggest that starting with an easy target domain does not necessarily lead to better performance. A harder target(larger domain shift) seems to be preferred for our method to obtain a better accuracy for future target domains. This is potentially due to the fact that features learned from a harder (larger domain shift) target tends to be more robust and more generalized than features learned with an easier target(small domain shift). While there can be a decrease given a scenario, overall, the results of different orders still perform better than the baselines. In addition, while the selection order can be important for good performance, in real-world applications, the order of targets is determined by the availability of data since we may not have access to all the targets at the same time. For the Wildtrack scenario, since there are six target domains, it means there are 720 permutations possible of the order of targets thus we selected a few orders based on domain shift (cosine distance) in Table \ref{tb:comp_cosine} in order to help our analysis. The chosen orders are: 1) small to large domain shift 2) Large to small domain shift 3) Alternate (alt.) between small and large shift starting with small shift 4) Alternate (alt.) between large and small shift starting with large shift 5) Our original order is actually another alternate (alt.) between large and small shift starting with a large shift.

\begin{table}[h]
\caption{Overall accuracy of MTDA-DTM on Wildtrack scenario with different target adaptation orders.}
\label{tb:comp_mtda_wildtrack_different_order}
\resizebox{\columnwidth}{!}{
\begin{tabular}{|l||r|r|r|r|r|r||r|}
\hline
\textbf{Backbone}: Resnet50                                                                                                                           & \textbf{C2}   & \textbf{C3}   & \textbf{C4}   & \textbf{C5}   & \textbf{C6}   & \textbf{C7}   & \textbf{Average} \\ \hline \hline
\multicolumn{8}{|l|}{\textbf{Train: Different Target Orders - Test: all targets}}       \\ \hline \hline
C1 $\xrightarrow{}$ C3 $\xrightarrow{}$ C4 $\xrightarrow{}$ C6 $\xrightarrow{}$ C5 $\xrightarrow{}$ C2 $\xrightarrow{}$ C7 (small $\xrightarrow{}$ large shift) & 49.4 & 57.7 & 54.2 & 56.3 & 39.1 & 68.8 & 54.2    \\ \hline
C1 $\xrightarrow{}$ C2 $\xrightarrow{}$ C7 $\xrightarrow{}$ C5 $\xrightarrow{}$ C6 $\xrightarrow{}$ C3 $\xrightarrow{}$ C4 (large $\xrightarrow{}$ small shift) & 48.4 & 58.4 & 51.6 & 58.7 & 40.3 & 68.4 & 54.3    \\ \hline
C1 $\xrightarrow{}$ C3 $\xrightarrow{}$ C2 $\xrightarrow{}$ C4 $\xrightarrow{}$ C7 $\xrightarrow{}$ C6 $\xrightarrow{}$ C5 (small/large shift alt.)              & 49.1 & 64.0 & 51.1 & 55.4 & 38.1 & 67.9 & 54.3   \\ \hline
C1 $\xrightarrow{}$ C2 $\xrightarrow{}$ C3 $\xrightarrow{}$ C7 $\xrightarrow{}$ C4 $\xrightarrow{}$ C5 $\xrightarrow{}$ C6 (large/small shift alt. 1)              & 49.7 & 65.3 & 53.9 & 57.7 & 40.4 & 68.0 & 55.8    \\ \hline
C1 $\xrightarrow{}$ C2 $\xrightarrow{}$ C3 $\xrightarrow{}$ C4 $\xrightarrow{}$ C5 $\xrightarrow{}$ C6 $\xrightarrow{}$ C7 (large/small shift alt. 2)              & 50.0 & 65.9 & 54.3 & 57.2 & 40.8 & 68.2 & 56.1    \\ \hline
\end{tabular}
}
\end{table}

Table \ref{tb:comp_mtda_wildtrack_different_order} gives us deeper insights compared to Table \ref{tb:comp_mtda_wcc_different_order} since it contains more target domains. Based on the results, we observe that, for this dataset, the orders that give the best results are the ones that start with large domain shift while alternating with smaller domain shift. This ordering makes sense since the detection model would always try to balance out between large domain shift and small domain shift thus allowing better domain-invariant features. As for starting with a target of large domains shift, this confirms the previous conclusion and this would mean a better starting weight for future target domains since adapting to large domain shift will allow the detector to have better domain-invariant features for the next target thus allowing an increase in accuracy. While these two factors, domain shift, and starting target, are the main factors that come out in our experiments. Other factors can potentially affect the final accuracy such as the difficulty of targets themselves or the number of classes.

\paragraph{Domain Shift Analysis}

In this experiment, we provide a visualization of the  domain shift in the problem of detection and incremental learning. First, we evaluate the domain shift using cosine distance on sources vs target features which is the most used in domain shift evaluation for classification. The features for the distance are computed using a detector trained only on the source to show the shift between source and target.

\begin{table}[h]
\centering
\caption{Cosine distance between source and target features on MTDA benchmarks.}
\label{tb:comp_cosine}
\resizebox{0.9\columnwidth}{!}{
\begin{tabular}{|l||l|l|l|l|l|l|}
\hline
\textbf{Targets} & \textbf{FoggyCityscape} & \textbf{RainCityscape} & \multicolumn{4}{l|}{} \\ \hline
\textbf{Source}: Cityscape     & 0.41       & 0.43  & \multicolumn{4}{l|}{} \\ \hline \hline
\textbf{Targets} & \textbf{Clipart} & \textbf{Watercolor} & \textbf{Comic} & \multicolumn{3}{l|}{} \\ \hline
\textbf{Source}: PascalVOC     & 0.67    & 0.66       & 0.67  & \multicolumn{3}{l|}{} \\ \hline \hline
\textbf{Targets} & \textbf{C2}      & \textbf{C3}         & \textbf{C4}    & \textbf{C5}    & \textbf{C6}    & \textbf{C7}    \\ \hline
\textbf{Source}: C1      & 0.66    & 0.63       & 0.63  & 0.65  & 0.64  & 0.66  \\ \hline
\end{tabular}
}
\end{table}

From Table \ref{tb:comp_cosine}, we can confirm that there is a domain shift between features of source and target domains at the global level. The problem with this in detection is that we cannot identify whether the domain shift comes from detection, i.e. region proposal, or it comes from the classification process. In order to see the domain shift at both detection and classification levels, we use the histogram of confidence score from both detection and classification on a source-only detector. For classification, we select the class "Person" for the classification confidence score between domains. For our histogram, a bin of 100 is used for detection since there are many bounding box proposals at the region proposal stage, the number of bins for classification is 10 for better visualization. For ease of visualization, we only show the shift between the source domain and one target domain.

\begin{figure*}[h!]
    \centering
    \includegraphics[width=0.9\textwidth]{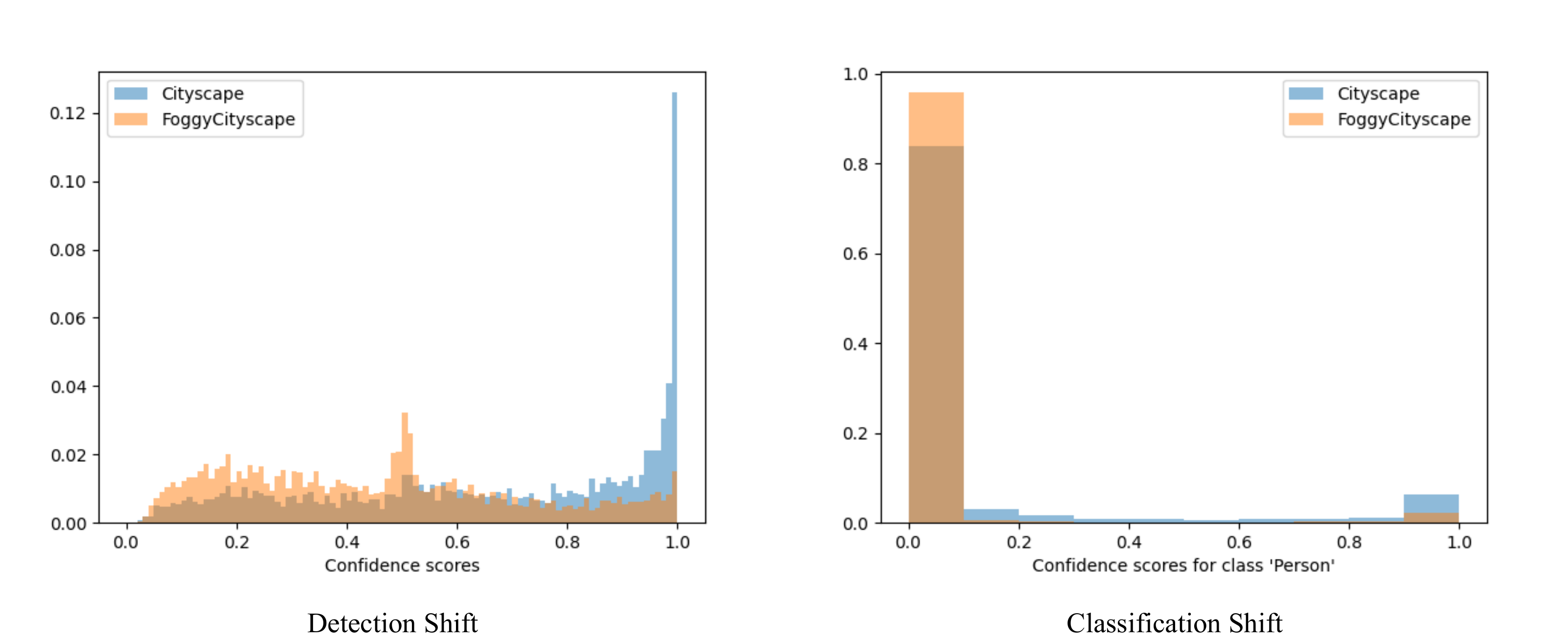}
    \caption{Visualization of shift in confidence scores on both detection and classification between Cityscape (source) and FoggyCityscape (target). The class used for classification shift is "Person".}
    \label{fig:hist_CS}
    \vspace{-3mm}
\end{figure*}

\begin{figure*}[h!]
    \centering
    \includegraphics[width=0.9\textwidth]{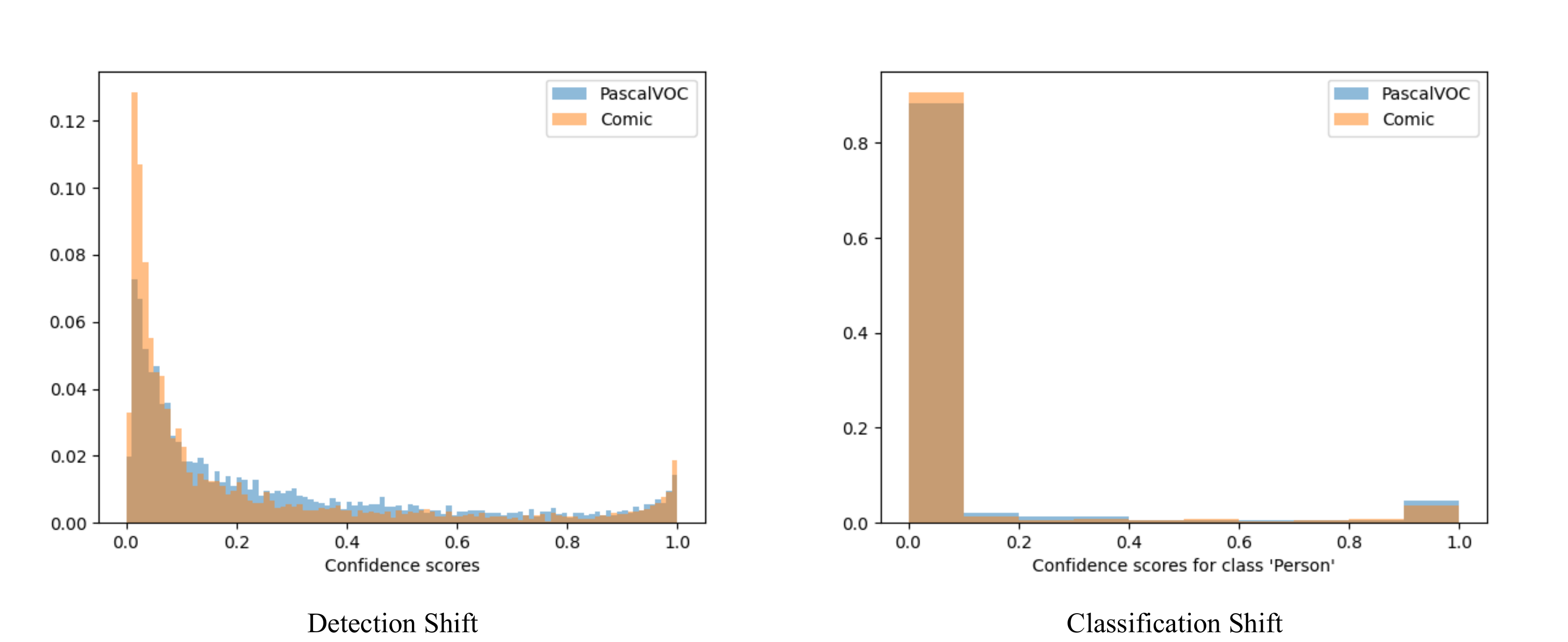}
    \caption{Visualization of shift in confidence scores on both detection and classification between PascalVOC (source) and Comic (target). The class used for classification shift is "Person".}
    \label{fig:hist_WCC}
    \vspace{-3mm}
\end{figure*}

\begin{figure*}[h!]
    \centering
    \includegraphics[width=0.9\textwidth]{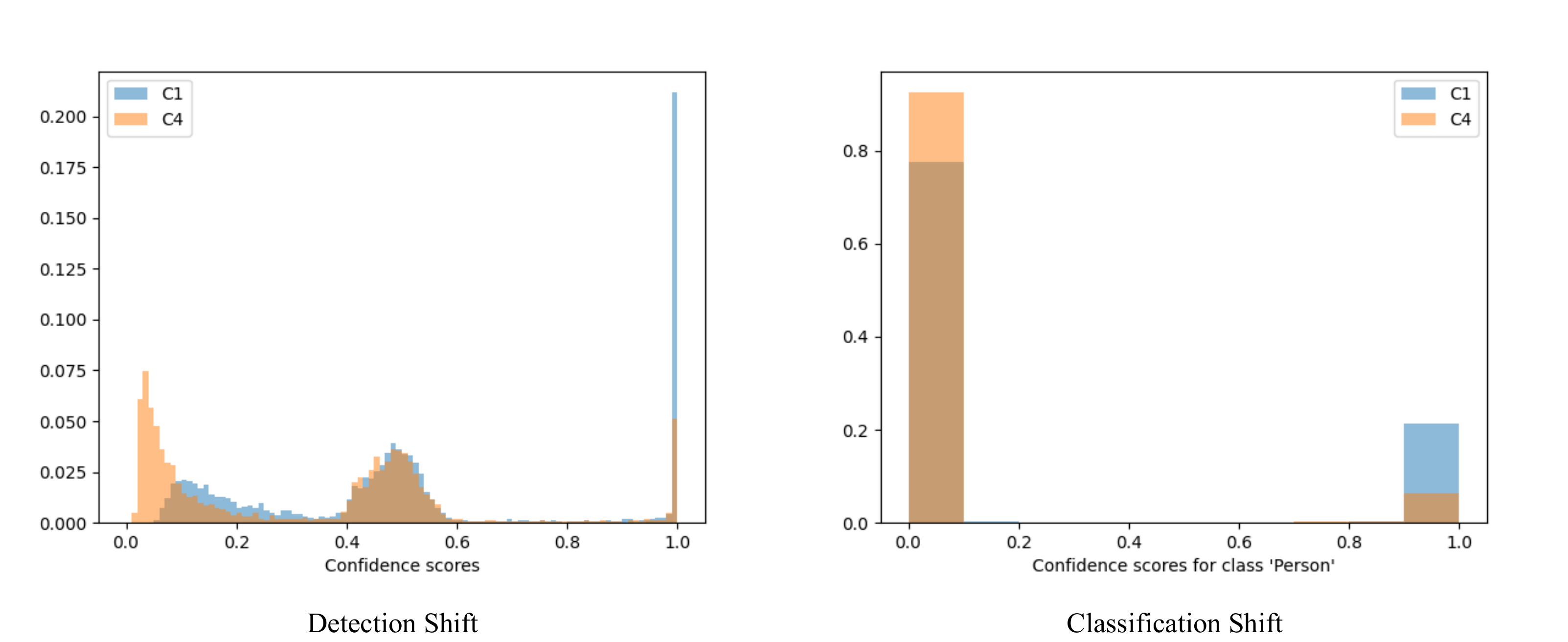}
    \caption{Visualization of shift in confidence scores on both detection and classification between Wildtrack (source) Camera 1 and Wildtrack Camera 4 (target). The class used for classification shift is "Person".}
    \label{fig:hist_WT}
    \vspace{-3mm}
\end{figure*}

From Figure \ref{fig:hist_CS}, we observe that the shift comes mostly from the detection task while there is only a slight shift on classification. This makes sense since Fog was added to Cityscape in an artificial way thus changing the detection problem without changing so much on the instances. In the Wildtrack problem, while the detection shift is still more pronounced albeit not as much as in Cityscape, the shift in classification is more pronounced. Finally, for the scenario of PascalVOC, we observe that there's not much change in confidence shift, this is confirmed by the results in Table \ref{tb:comp_mtda_voc_clipart_2_watercolor_eval_clipart_watercolor} where the Source Only has similar performance compared to state-of-the-art like \cite{wei2020_inc_mtda} or UFT baseline.

\vspace{-3mm}
\subsection{Complexity Analysis:}

Object detectors can require a high computational complexity, especially region-based detectors like Faster R-CNN. While the inference complexity is important, in this work, we are interested in the domain adaptation complexity since it's very important for a technique to have an efficient training so that it can be easily adopted in real-world applications. In addition, since our method and existing baselines do not add extra complexity, the measured complexity would be the same. The (memory and computation) complexity for adaptation between our method and the current state-of-the-art is analyzed. For incremental domain adaptation, the method in Incr. MTDA KD \cite{wei2020_inc_mtda} uses a duplicated detector and distillation to avoid catastrophic forgetting. This approach, however, introduces a significant overhead. Theoretically, a detector that uses $C_m$ amount of memory and $C_f$ amount of computation, for an incremental domain adaptation with a duplicated detection model it would take $2\times C_M$ and $2 \times C_f$ during each incremental step. Using MT-MTDA \cite{mt-mtda}, (state-of-the-art in image classification), has one teacher model per target domain, as well as one single student detector. Assuming that backbone of teachers and student is ResNet50, where $n$ is the number of target domains. Then this approach requires $n\times C_M$ and $n \times C_f$. In the following Table \ref{tb:comp_complexity}, we show the complexity of these methods, Incr. MTDA KD\cite{wei2020_inc_mtda} and MT-MTDA \cite{mt-mtda} with an image RGB of size $3\times1200\times600$ with $3$ and $7$ target domains. The complexity is measured over one iteration (one forward pass), for a whole training, the measure can be obtained by multiplying these measurements with the total number of iterations.

\begin{table}[h]
\centering
\caption{Comparison of memory complexity (number of detection model parameters) and time complexity (FLOPS for training with $3$ or $7$ target domains) given input images of $1200\times600$ for one forward pass.}
\label{tb:comp_complexity}
\resizebox{\textwidth}{!}{
\begin{tabular}{|l||r|r|r|r||r|r|}
\hline
\textbf{Backbone CNN}    & \multicolumn{4}{c||}{\textbf{Resnet50}} & \multicolumn{2}{c|}{\textbf{Resnet101}} \\ \hline 
\textbf{No. of target domains} & \multicolumn{2}{c|}{\textbf{3}}       & \multicolumn{2}{c||}{\textbf{7}}       & \multicolumn{2}{c|}{\textbf{7}}         \\ \hline 
\textbf{Methods} & \textbf{\# para. (M)} & \textbf{GFLOPS} & \textbf{\# para. (M)} & \textbf{GFLOPS} & \textbf{\# para. (M)}  & \textbf{GFLOPS}  \\ \hline \hline
Incr. MTDA KD \cite{wei2020_inc_mtda}   & 74 & 322  & 74 & 322  & 112  & 430   \\ \hline
MT-MTDA     & 148   & 644  & 259 & 1127 & 392   & 1505  \\ \hline
MTDA-DTM (ours) & \textbf{38}  & \textbf{162}  & \textbf{38}  & \textbf{162}  & \textbf{56}    & \textbf{216}   \\ \hline \hline
UFT & 37  & 161  & 37  & 161  & 55    & 215   \\ \hline
\end{tabular}
}
\end{table}

From Table \ref{tb:comp_complexity}, MT-MTDA \cite{mt-mtda} requires the most in term of resources and is clearly not scalable for more target domains. While it improves over MT-MTDA, Incr. MTDA KD \cite{wei2020_inc_mtda}, it is still $2$ more complex than MTDA-DTM. Not only can MTDA-DTM achieve better performance in terms of accuracy, it only requires an overhead of $1G$ of total FLOPS and $1.5k$ parameters regardless of any architecture.

\begin{figure*}[htbp]
    \centering
    \includegraphics[width=0.75\textwidth]{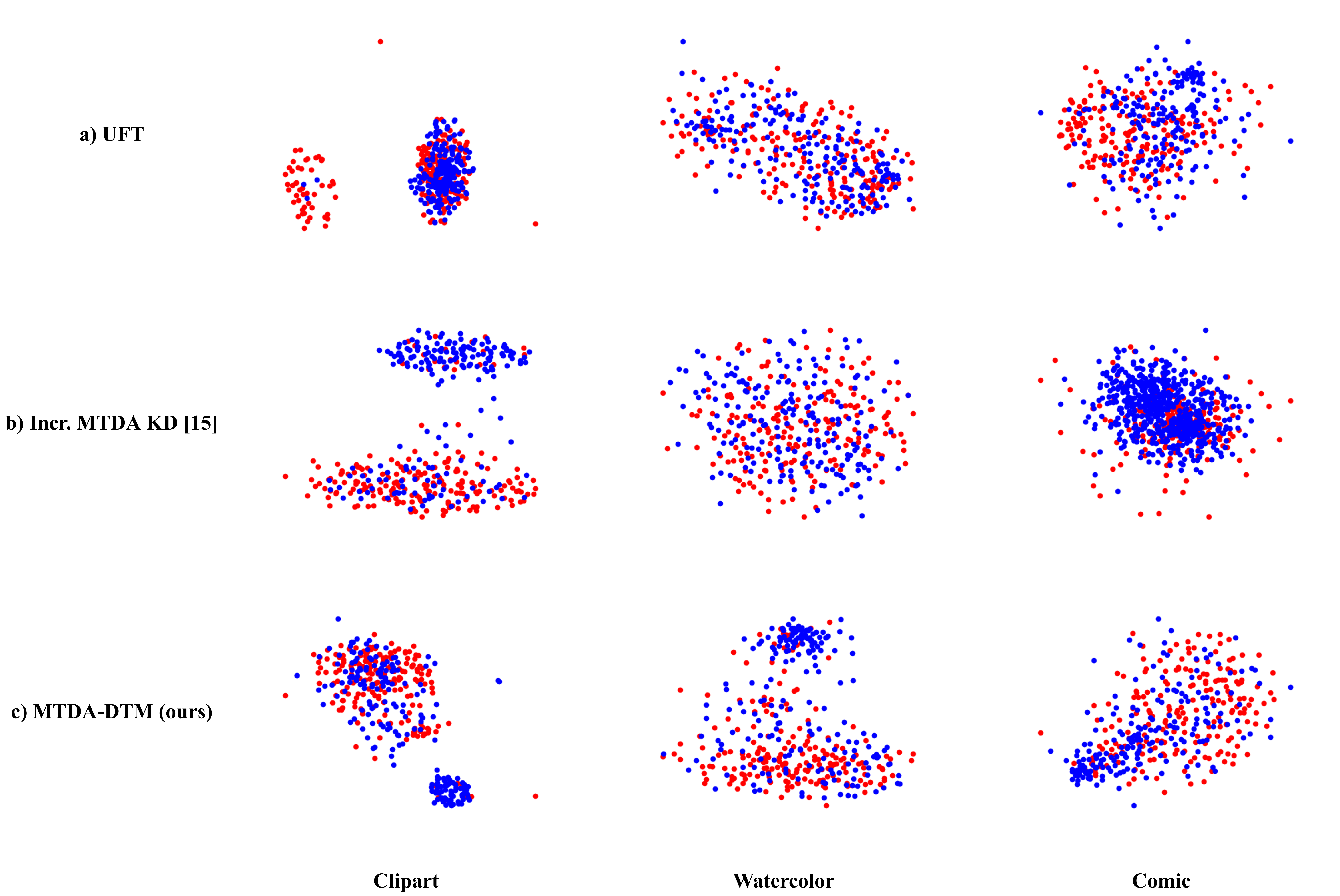}
    \caption{UMAP feature visualization of (a) UFT, (b) Incr. MTDA KD of Wei et al. \cite{wei2020_inc_mtda} and (c) MTDA-DTM (ours) after Pascal (\textbf{$S$})  $\xrightarrow{}$ Clipart (\textbf{$T_1$})  $\xrightarrow{}$ Watercolor (\textbf{$T_2$})  $\xrightarrow{}$ Comic (\textbf{$T_3$})   on each target domains, with source (red) and target(blue). Best viewed in color.}
    \label{fig:umap_mtda}
    \vspace{-3mm}
\end{figure*}

\vspace{-3mm}
\subsection{Feature Visualization:}

The UMAP method was employed for feature visualization of baselines such as UFT, Incr. MTDA KD \cite{wei2020_inc_mtda} on 200 samples of deep features of each target domains of PascalVOC scenario using a detection model after Pascal (\textbf{$S$})  $\xrightarrow{}$ Clipart (\textbf{$T_1$})  $\xrightarrow{}$ Watercolor (\textbf{$T_2$})  $\xrightarrow{}$ Comic (\textbf{$T_3$}).

From Fig. \ref{fig:umap_mtda}, we can see that all detectors can mix the data between source and latest domain relatively well. One notable difference can be observed at the Clipart target domain, where target features of "a) UFT" and "b) Incr. MTDA KD of Wei et al. \cite{wei2020_inc_mtda}" are further and have less overlap with source features. Since the source domain is where detectors are supposed to excel given they have access to labels, it makes sense that our MTDA-DTM performs better than others given that target features are closer to the source and there are more target features overlapping with source features.

\vspace{-3mm}
\section{Conclusion}

In this paper, a novel approach is introduced for efficient MTDA of object detectors that can achieve a high level of accuracy on multiple target domains without labels. The proposed approach, MTDA-DTM, is suitable for real-world applications since it allows the detector to be incrementally updated without degrading performance on data from previously-learned target domains. This is achieved using a Domain Transfer Module (DTM), that takes advantage of domain discriminators and does not need data from previously-learned target domains. In addition, DTM only requires a small memory and computational overhead compared to existing methods. 

Results indicate that our proposed method outperforms the state-of-art methods, and even ideal baselines, especially when dealing a growing number of target domains. In some cases, our MTDA-DTM can provide comparable performance with supervised training. One of our ablation studies also shows that MTDA-DTM outperforms the state-of-the-art in MTDA that uses multiple teachers to distill to a single student. Therefore, adapting to a target one at a time can be very beneficial for MTDA. In addition to showing our performance on MTDA across dataset benchmarks, we also show results of our detector in real-world application settings like multi-camera adaptation for object detection. Results indicate that our detection model still improves upon baseline and state-of-the-art models. Finally, our method generalizes well on adapted domains, and is robust to unseen target domains, and thus  more suitable for real-world applications. \textcolor{black}{Our proposed MTDA-DTM depends on the order of targets selected for incremental leaning. Future work should be focused on techniques that are independent of the training order, or that determine the optimal order based on the properties of target domains.}


\begin{small}
\setstretch{0.9}

\bibliography{egbib}

\begin{thebibliography}{10}
\expandafter\ifx\csname url\endcsname\relax
  \def\url#1{\texttt{#1}}\fi
\expandafter\ifx\csname urlprefix\endcsname\relax\def\urlprefix{URL }\fi
\expandafter\ifx\csname href\endcsname\relax
  \def\href#1#2{#2} \def\path#1{#1}\fi

\bibitem{frcnn}
S.~Ren, K.~He, R.~Girshick, J.~Sun, Faster r-cnn: Towards real-time object
  detection with region proposal networks, in: NeurIPS, 2015.

\bibitem{zhao2019object}
Z.-Q. Zhao, P.~Zheng, S.-t. Xu, X.~Wu, Object detection with deep learning: A
  review, IEEE Trans. on Neural Networks and Learning Systems 30~(11) (2019)
  3212--3232.

\bibitem{efficientdet}
M.~Tan, R.~Pang, Q.~V. Le, Efficientdet: Scalable and efficient object
  detection, in: CVPR 2020.

\bibitem{FRCNN_DA}
Y.~Chen, W.~Li, C.~Sakaridis, D.~Dai, L.~Van~Gool, Domain adaptive faster r-cnn
  for object detection in the wild, in: CVPR, 2018.

\bibitem{GRL}
Y.~Ganin, V.~Lempitsky, Unsupervised domain adaptation by backpropagation, in:
  ICML 2015.

\bibitem{HTCN}
C.~Chen, Z.~Zheng, X.~Ding, Y.~Huang, Q.~Dou, Harmonizing transferability and
  discriminability for adapting object detectors, in: CVPR, 2020.

\bibitem{Progressive_DA_FRCNN}
H.-K. Hsu, C.-H. Yao, Y.-H. Tsai, W.-C. Hung, H.-Y. Tseng, M.~Singh, M.-H.
  Yang, Progressive domain adaptation for object detection, in: WACV, 2020.

\bibitem{CycleGAN2017}
J.-Y. Zhu, T.~Park, P.~Isola, A.~A. Efros, Unpaired image-to-image translation
  using cycle-consistent adversarial networkss, in: ICCV, 2017.

\bibitem{mt-mtda}
L.~T. Nguyen-Meidine, A.~Belal, M.~Kiran, J.~Dolz, L.-A. Blais-Morin,
  E.~Granger, Unsupervised multi-target domain adaptation through knowledge
  distillation, in: WACV, 2021.

\bibitem{wei2020_inc_mtda}
X.~Wei, S.~Liu, Y.~Xiang, Z.~Duan, C.~Zhao, Y.~Lu, Incremental learning based
  multi-domain adaptation for object detection, Knowledge-Based Systems 210
  (2020) p. 106420.

\bibitem{BlendMTDA}
Z.~Chen, J.~Zhuang, X.~Liang, L.~Lin, Blending-target domain adaptation by
  adversarial meta-adaptation networks, in: CVPR, 2019.

\bibitem{dada}
X.~Peng, Z.~Huang, X.~Sun, K.~Saenko, Domain agnostic learning with
  disentangled representations, in: ICML, 2019.

\bibitem{IJCNN_KD_UDA}
L.~T. Nguyen-Meidine, E.~Granger, M.~Kiran, J.~Dolz, L.-A. Blais-Morin, Joint
  progressive knowledge distillation and unsupervised domain adaptation, IJCNN
  2020.

\bibitem{explaining_harnessing_adversarial_ex}
I.~Goodfellow, J.~Shlens, C.~Szegedy, Explaining and harnessing adversarial
  examples, in: ICLR, 2015.

\bibitem{GAN}
I.~Goodfellow, J.~Pouget-Abadie, M.~Mirza, B.~Xu, D.~Warde-Farley, S.~Ozair,
  A.~Courville, Y.~Bengio, Generative adversarial nets, in: NeurIPS 2014.

\bibitem{PascalVOC}
M.~Everingham, L.~Van~Gool, C.~K.~I. Williams, J.~Winn, A.~Zisserman, The
  pascal visual object classes (voc) challenge, Int'l Journal of Computer
  Vision 88~(2) (2010) 303--338.

\bibitem{ClipWaterComic}
N.~Inoue, R.~Furuta, T.~Yamasaki, K.~Aizawa, Cross-domain weakly-supervised
  object detection through progressive domain adaptation, in: CVPR, 2018.

\bibitem{cityscape}
M.~Cordts, M.~Omran, S.~Ramos, T.~Rehfeld, M.~Enzweiler, R.~Benenson,
  U.~Franke, S.~Roth, B.~Schiele, The cityscapes dataset for semantic urban
  scene understanding, in: CVPR 2016.

\bibitem{FoggyCityscape}
C.~Sakaridis, D.~Dai, L.~Van~Gool, Semantic foggy scene understanding with
  synthetic data, Int'l Journal of Computer Vision ( 126~(9) (2018) 973--992.

\bibitem{RainCityscape}
X.~Hu, C.-W. Fu, L.~Zhu, P.-A. Heng, Depth-attentional features for
  single-image rain removal, in: CVPR 2019.

\bibitem{Wildtrack}
T.~Chavdarova, P.~Baqué, S.~Bouquet, A.~Maksai, C.~Jose, T.~Bagautdinov,
  L.~Lettry, P.~Fua, L.~Van~Gool, F.~Fleuret, Wildtrack: A multi-camera hd
  dataset for dense unscripted pedestrian detection, in: CVPR, 2018.

\bibitem{DA_PR_GAN}
W.~Chen, H.~Hu, Generative attention adversarial classification network for
  unsupervised domain adaptation, Pattern Recognition 107 (2020) p. 107440.

\bibitem{DA_PR_seg_transla}
R.~Li, W.~Cao, Q.~Jiao, S.~Wu, H.-S. Wong, Simplified unsupervised image
  translation for semantic segmentation adaptation, Pattern Recognition 105
  (2020) p. 107343.

\bibitem{DA_Pr_Seg_scale_inv}
D.~Guan, J.~Huang, S.~Lu, A.~Xiao, Scale variance minimization for unsupervised
  domain adaptation in image segmentation, Pattern Recognition 112 (2021) p.
  107764.

\bibitem{Instance-Invariant_DA_OD}
A.~Wu, Y.~Han, L.~Zhu, Y.~Yang, Instance-invariant domain adaptive object
  detection via progressive disentanglement, IEEE Transactions on Pattern
  Analysis and Machine Intelligence (2021) 1--1\href
  {http://dx.doi.org/10.1109/TPAMI.2021.3060446}
  {\path{doi:10.1109/TPAMI.2021.3060446}}.

\bibitem{saito2019strong}
K.~Saito, Y.~Ushiku, T.~Harada, K.~Saenko, Strong-weak distribution alignment
  for adaptive object detection, in: CVPR, 2019.

\bibitem{uda_od_cross_domain_aligment}
X.~Zhu, J.~Pang, C.~Yang, J.~Shi, D.~Lin, Adapting object detectors via
  selective cross-domain alignment, in: CVPR, 2019.

\bibitem{multiadversarial_frcnn}
Z.~He, L.~Zhang, Multi-adversarial faster-rcnn for unrestricted object
  detection, in: ICCV 2019.

\bibitem{diversify_frcnn}
T.~Kim, M.~Jeong, S.~Kim, S.~Choi, C.~Kim, Diversify and match: A domain
  adaptive representation learning paradigm for object detection, in: CVPR,
  2019.

\bibitem{Uncertainty-Aware_DA_OD}
D.~Guan, J.~Huang, A.~Xiao, S.~Lu, Y.~Cao, Uncertainty-aware unsupervised
  domain adaptation in object detection, IEEE Transactions on Multimedia (2021)
  1--1\href {http://dx.doi.org/10.1109/TMM.2021.3082687}
  {\path{doi:10.1109/TMM.2021.3082687}}.

\bibitem{Self_guided_DA_Det}
C.~Zhang, Z.~Li, J.~Liu, P.~Peng, Q.~Ye, S.~Lu, T.~Huang, Y.~Tian, Self-guided
  adaptation: Progressive representation alignment for domain adaptive object
  detection, IEEE Transactions on Multimedia (2021) 1--1\href
  {http://dx.doi.org/10.1109/TMM.2021.3078141}
  {\path{doi:10.1109/TMM.2021.3078141}}.

\bibitem{MTDA_Theoric}
B.~Gholami, P.~Sahu, O.~Rudovic, K.~Bousmalis, V.~Pavlovic, Unsupervised
  multi-target domain adaptation: An information theoretic approach, IEEE
  Trans. on Image Processing 29 (2020) 3993--4002.

\bibitem{MTDA_Segmentation}
T.~Isobe, X.~Jia, S.~Chen, J.~He, Y.~Shi, J.~Liu, H.~Lu, S.~Wang, Multi-target
  domain adaptation with collaborative consistency learning, in: Proceedings of
  the IEEE/CVF Conference on Computer Vision and Pattern Recognition (CVPR),
  2021, pp. 8187--8196.

\bibitem{EWC}
J.~Kirkpatrick, R.~Pascanu, N.~Rabinowitz, J.~Veness, G.~Desjardins, A.~A.
  Rusu, K.~Milan, J.~Quan, T.~Ramalho, A.~Grabska-Barwinska, D.~Hassabis,
  C.~Clopath, D.~Kumaran, R.~Hadsell, Overcoming catastrophic forgetting in
  neural networks, Proceedings of the National Academy of Sciences 114~(13)
  (2017) 3521--3526.

\bibitem{IL_SI}
F.~Zenke, B.~Poole, S.~Ganguli, Continual learning through synaptic
  intelligence, in: ICML, 2017.

\bibitem{IL_Lwf}
Z.~Li, D.~Hoiem, Learning without forgetting, IEEE Trans. on Pattern Analysis
  and Machine Intelligence 40~(12) (2018) 2935--2947.

\bibitem{Hou_2019_CVPR}
S.~Hou, X.~Pan, C.~C. Loy, Z.~Wang, D.~Lin, Learning a unified classifier
  incrementally via rebalancing, in: CVPR, 2019.

\bibitem{IL_PC}
J.~Schwarz, J.~Luketina, W.~M. Czarnecki, A.~Grabska-Barwinska, Y.~W. Teh,
  R.~Pascanu, R.~Hadsell, Progress \& compress: A scalable framework for
  continual learning, ICML 2018.

\bibitem{IL_WR_KD}
B.~Zhao, X.~Xiao, G.~Gan, B.~Zhang, S.~Xia, Maintaining discrimination and
  fairness in class incremental learning, in: CVPR, 2020.

\bibitem{MetaER}
M.~Riemer, I.~Cases, R.~Ajemian, M.~Liu, I.~Rish, Y.~Tu, G.~Tesauro, Learning
  to learn without forgetting by maximizing transfer and minimizing
  interference, in: ICLR 2019.

\bibitem{IL_RH_Vae}
H.~Shin, J.~K. Lee, J.~Kim, J.~Kim, Continual learning with deep generative
  replay (2017), in: NeurIPS 2017.

\bibitem{IL_AR_1}
R.~Aljundi, P.~Chakravarty, T.~Tuytelaars, Expert gate: Lifelong learning with
  a network of experts, in: CVPR, 2016.

\bibitem{IL_AR_2}
S.~S. Sarwar, A.~Ankit, K.~Roy, Incremental learning in deep convolutional
  neural networks using partial network sharing, IEEE Access 8 (2020)
  4615--4628.

\bibitem{darER}
P.~Buzzega, M.~Boschini, A.~Porrello, D.~Abati, S.~Calderara, Dark experience
  for general continual learning: a strong, simple baseline, in: NeurIPS 2020.

\bibitem{Inc_few_shot_sup}
J.-M. Perez-Rua, X.~Zhu, T.~M. Hospedales, T.~Xiang, Incremental few-shot
  object detection, in: CVPR, 2020.

\bibitem{Inc_OD_2017_ICCV}
K.~Shmelkov, C.~Schmid, K.~Alahari, Incremental learning of object detectors
  without catastrophic forgetting, in: ICCV, 2017.

\bibitem{mtl_self_supervised_od}
W.~Lee, J.~Na, G.~Kim, Multi-task self-supervised object detection via
  recycling of bounding box annotations, in: CVPR 2019.

\bibitem{mtl_spotnet}
H.~Perreault, G.-A. Bilodeau, N.~Saunier, M.~Héritier, Spotnet: Self-attention
  multi-task network for object detection (2020), in: Conference on Computer
  and Robot Vision (CRV), 2020.

\bibitem{multi_task_inc}
X.~Liu, H.~Yang, A.~Ravichandran, R.~Bhotika, S.~Soatto, Multi-task incremental
  learning for object detection (2020), CoRR abs/2002.05347.

\bibitem{FocalLoss}
T.-Y. Lin, P.~Goyal, R.~Girshick, K.~He, P.~Dollar, Focal loss for dense object
  detection, in: ICCV 2017.

\bibitem{Overhaul}
B.~Heo, J.~Kim, S.~Yun, H.~Park, N.~Kwak, J.~Y. Choi, A comprehensive overhaul
  of feature distillation, in: ICCV, 2019.

\bibitem{umap}
L.~McInnes, J.~Healy, N.~Saul, L.~Großberger, Umap: Uniform manifold
  approximation and projection, Journal of Open Source Software 3~(29) (2018)
  861.

\bibitem{TSNE}
L.~van~der Maaten, G.~Hinton, Visualizing data using t-sne, Journal of Machine
  Learning Research 9~(86) (2008) 2579--2605.

\end{thebibliography}
\end{small}

\newpage
\appendix
\clearpage
{ \centering \Large \textbf{Appendices}}

\section{Hyper-Parameter Values}

Table \ref{tb:hyper-parameters} shows hyper-parameters values selected for our incremental domain adaptation and DTM module. The HTCN \cite{HTCN} model is used as the baseline for our MTDA approach, and specific details on HTCN can be found in their original paper \cite{HTCN}. Hyper-parameters for the first domain adaptation are therefore based on HTCN. For 2-7 incremental domain adaptation step, the number of iterations and the number of epochs remain the same, while others and hyper-parameters for DTM were chosen using a separate hold-out validation process.

\begin{table}[hbp]
\large
\caption{Value of hyper-parameters selected for our proposed model.}
\label{tb:hyper-parameters}
\centering
\resizebox{\textwidth}{!}{
\begin{tabular}{|l||c|c|c|c|c|c|}
	\hline
\textbf{Hyper-parameters}  & \multicolumn{2}{|c|}{\textbf{Wildtrack}}  & \multicolumn{2}{|c|}{\textbf{Cityscape}}  & \multicolumn{2}{|c|}{\textbf{PascalVOC}}    \\
incremental adaptation step  &  1  &  2-7 &  1 &  2 &  1 &  2-3 \\\hline \hline
	\multicolumn{7}{|l|}{\textbf{Incremental Domain Adaptation} }                                                                                               \\ \hline \hline
	learning rate  & 0.001    & 0.0001        & 0.001          & 0.0001        & 0.001        & 0.0001         \\ \hline
	momentum       & \multicolumn{6}{|c|}{0.9}                          \\ \hline
	$\alpha$       & -                        & 1                            & -                         & 0.1                       & -                         & 1                             \\ \hline
	number of iterations  per epoch & \multicolumn{2}{|c|}{800}            & \multicolumn{4}{|c|}{10000} \\ \hline
	number of epochs  &  \multicolumn{6}{|c|}{7}                       \\ \hline
	learning rate after 5 epochs & 0.0001                   & 0.00001                      & 0.0001                    & 0.00001                   & 0.0001                    & 0.00001                       \\ \hline
	$\lambda$ of HTCN  & \multicolumn{6}{|c|}{1}                             \\ \hline \hline
	\multicolumn{7}{|l|}{\textbf{DTM}   }                                                                                                      \\ \hline \hline
	learning rate           & \multicolumn{6}{|c|}{0.01} \\ \hline
	momentum                & \multicolumn{6}{|c|}{0.9} \\ \hline
	number of iterations   & \multicolumn{2}{|c|}{800}            & \multicolumn{4}{|c|}{10000} \\ \hline
\end{tabular}
}
\end{table}

\section{Importance of $\alpha$}
In this experiment, the importance of $\alpha$ used in Equation \ref{eq:ida} is empirically validated. For this experiment, we used the Cityscape scenario \textcolor{black}{and the PascalVOC $\xrightarrow{}$ Clipart $\xrightarrow{}$ Watercolor scenario as these two scenarios have used different $\alpha$}
\newline

\begin{table}[h!]
\caption{Average Precision of MTDA-DTM when varying the $\alpha$ parameter on Cityscape scenario.}
\label{tb:alpha_abl}
\centering
\resizebox{0.8\columnwidth}{!}{
\begin{tabular}{|l||r|r|r|r|r|r|r|r||r|}
\hline
\textbf{Backbone}: VGG16  & \multicolumn{9}{c|}{\textbf{Accuracy}}           \\ 
\textbf{$\alpha$ value}          & \textbf{bus}  & \textbf{bicycle} & \textbf{car}   & \textbf{m.cycle} & \textbf{person} & \textbf{rider} & \textbf{train} & \textbf{truck} & \textbf{mAP}      \\ \hline \hline
\multicolumn{10}{|l|}{\textbf{Train: Cityscape (\textbf{$S$})  $\xrightarrow{}$ FoggyCityscape (\textbf{$T_1$})  $\xrightarrow{}$ RainCityscape (\textbf{$T_2$}) - Test: all targets}}                                                         \\ \hline 

0.1       & \textbf{66.0}   & 34.6   & 48.4 & 18.2       & 29.3  & 53.5  & 35.4 & \textbf{27.6} & \textbf{39.1} \\ \hline
0.2       & \textbf{66.0} & 34.4    & 48.4  & 18.0       & 29.0   & 53.9  & 32.9  & 27.5  & 38.8  \\ \hline
0.3       & 65.2 & \textbf{35.8}    & \textbf{48.7}  & 21.0         & 29.3   & 53.1  & 32.5  & 26.9  & 39.0  \\ \hline
0.4       & 65.1 & 34.6    & 48.2  & 20.1       & 29.2   & 53.5  & 34.9  & 26.1  & 38.9 \\ \hline
0.5       & 64.0   & 33.9    & 48.6  & 22.0         & 28.8   & \textbf{54.1}  & 34.7  & 26.2  & 39.0  \\ \hline
0.6       & 64.6 & 34.2    & 48.3  & 20.4       & 29.1   & 52.6  & 34.3  & 26.5  & 38.7    \\ \hline
0.7       & 64.2 & 34.7    & 47.8  & \textbf{24.0}         & \textbf{30.0}     & 53.4  & 32.1  & 24.2  & 38.8     \\ \hline
0.8       & 64.3 & 33.6    & 48.3  & 20.4       & 28.9   & 53.1  & \textbf{35.8}  & 24.9  & 38.6  \\ \hline
0.9       & 62.3 & 34.5    & 48.5  & 22.3       & 29.0     & 53.5  & 32.6  & 25.5  & 38.5   \\ \hline
1.0         & 65.3 & 33.9    & 48.2  & 19.9       & 29.2   & 53.0    & 34.7  & 22.5  & 38.3  \\ \hline
\end{tabular}
}
\end{table}

\vspace{-3mm}

\begin{table}[h!]
\color{black}
\centering
\caption{Average Precision of the proposed MTDA-DTM, baselines and state-of-the-art models for MTDA on PascalVOC $\xrightarrow{}$ Clipart $\xrightarrow{}$ Watercolor scenario.}
\label{tb:alpha_abl_2}
\resizebox{0.8\textwidth}{!}{
\begin{tabular}{|l||r|r|r|r|r|r||r|}
\hline
\textbf{Backbone}: Resnet50 \ \ \ & \multicolumn{7}{c|}{\textbf{Accuracy}}       \\ 
 \textbf{$\alpha$ value}  & \textbf{bicycle}                                                                        & \textbf{bird}  & \textbf{car}  & \textbf{cat}  & \textbf{dog}  & \textbf{person} & \textbf{mAP}  \\ \hline \hline
\multicolumn{8}{|l|}{\textbf{Train: PascalVOC (\textbf{$S$}) $\xrightarrow{}$ Clipart (\textbf{$T_1$}) $\xrightarrow{}$ Watercolor (\textbf{$T_2$}) - Test: all targets}}  \\ \hline \hline
0.1  & 56.7                                                     & 34.6   & 37.4  & 13.3   & 24.4     & 62.3    & 38.1  \\ \hline

0.2         & 58.2                                                     & 37.3   & 37.9   & 16.7  & 19.2   & 62.6    & 38.5  \\ \hline
0.3      & 57.7                                                     & 38.4  & 39.1  & 18.2  & \textbf{19.8}  & 62.7  & 39.3 \\ \hline
0.4         & 56.2                                                   & 38.3  & 40.6  & 14.6  & 24.0  & 61.8   & 39.1 \\ \hline 
0.5       & 53.0                                                   & \textbf{39.0} & 35.5  & 15.7    & 28.2 & 63.5    & 39.1 \\ \hline 
0.6    & 46.6                                                    & 38.2  & 37.3   & 19.8  & 23.2  & 62.7   & 39.1       \\ \hline
0.7        & 52.1                                                      & 37.0  & 38.5    & 19.4  & 26.1  & 63.1    & 39.4  \\\hline
0.8        & 52.6                                                   & 35.5  & 40.8   & 15.5  & \textbf{29.4}  & \textbf{64.1}    & 39.6  \\\hline
0.9        & 52.8                                                     & 37.2 & \textbf{41.4}   & 18.7  & 24.6  & \textbf{64.1}    & \textbf{39.8}  \\\hline
1        & \textbf{62.8}                                                    & 35.6 & \textbf{38.1} & 14.3 & 24.1 & 63.7   & \textbf{39.8} \\\hline

\end{tabular}
}
\vspace{-3mm}
\end{table}

Table \ref{tb:alpha_abl} shows that the further we increase the value of $\alpha$, the worse the performance drops, as seen in the case of $\alpha=1.0$. From the same Table \ref{tb:alpha_abl}, we see that our training can be relatively stable. There is no significant difference between variations of $\alpha$, the biggest difference is of $0.8$ in mAP between $\alpha=0.1$ and $\alpha=1$ and the difference between two sequential values of $\alpha$ is mostly around $0.2$, which means that our training is quite stable overall. \textcolor{black}{From the Table \ref{tb:alpha_abl_2}, we observe that our hypothesis about the value of $\alpha$ is confirmed. Our results increase when $\alpha$ is closer to $1$ as this scenario has a bigger shift that Cityscape.}

\section{Architecture of DTM}

\textcolor{black}{
In this experiment, we vary the architecture of DTM in order to evaluate its impact on performance. We explore 3 variations of DTM: 1) A wider DTM with 2 layers 2) A DTM with 4 layers 3) A DTM with 6 layers. These different architectures are described in Table \ref{tb:DTM_arch}.
}

\begin{table}[h!]
\color{black}
\caption{Average Precision of MTDA-DTM with different architecture of DTM on Cityscape scenario.}
\label{tb:DTM_arch}
\centering
\resizebox{0.8\columnwidth}{!}{
	\begin{tabular}{|l||lllc|}
		\hline
		\textbf{Architecture}    & \multicolumn{1}{c|}{\textbf{Original DTM}}                         & \multicolumn{1}{c|}{\textbf{Wider DTM}}                            & \multicolumn{1}{c|}{\textbf{4 Layers DTM}}                           & \textbf{6 Layers DTM}                           \\ \hline \hline
		& \multicolumn{4}{c|}{\textbf{Tensor dimension}: output channels $\times$ input channels $\times$ height $\times$ width}                                                                       \\ \hline
		1st conv. layer & \multicolumn{1}{c|}{256 $\times$ 3 $\times$ 1 $\times$ 1} & \multicolumn{1}{c|}{512 $\times$ 3 $\times$ 1 $\times$ 1} & \multicolumn{1}{c|}{256 $\times$ 3 $\times$ 1 $\times$ 1}   & 64 $\times$ 3 $\times$ 1 $\times$ 1    \\ \hline
		ReLU            & \multicolumn{1}{c|}{\checkmark}                                     & \multicolumn{1}{c|}{\checkmark}                                     & \multicolumn{1}{c|}{\checkmark}                                      & \multicolumn{1}{c|}{\checkmark}                   \\ \hline
		2nd conv. layer & \multicolumn{1}{c|}{3 x 256 $\times$ 1 $\times$ 1}        & \multicolumn{1}{c|}{3 $\times$ 512 $\times$ 1 $\times$ 1} & \multicolumn{1}{c|}{512 $\times$ 256 $\times$ 1 $\times$ 1} & 128 $\times$ 64 $\times$ 1 $\times$ 1  \\ \hline
		ReLU            & \multicolumn{1}{l|}{}                                     & \multicolumn{1}{l|}{}                                     & \multicolumn{1}{c|}{\checkmark}                                        &  \multicolumn{1}{c|}{\checkmark}                                        \\ \hline
		3rd conv. layer & \multicolumn{1}{l|}{}                                     & \multicolumn{1}{l|}{}                                     & \multicolumn{1}{c|}{256 $\times$ 512 $\times$ 1 $\times$ 1} & 256 $\times$ 128 $\times$ 1 $\times$ 1 \\ \hline
		ReLU            & \multicolumn{1}{l|}{}                                     & \multicolumn{1}{l|}{}                                     & \multicolumn{1}{c|}{\checkmark}                                       & \multicolumn{1}{c|}{\checkmark}                  \\ \hline
		4th conv. layer & \multicolumn{1}{l|}{}                                     & \multicolumn{1}{l|}{}                                     & \multicolumn{1}{c|}{3 $\times$ 256 $\times$ 1 $\times$ 1}   & 128 $\times$ 256 $\times$ 1 $\times$ 1 \\ \hline
		ReLU            & \multicolumn{1}{l|}{}                                     & \multicolumn{1}{l|}{}                                     & \multicolumn{1}{l|}{}                                       & \multicolumn{1}{c|}{\checkmark}                  \\ \hline
		3rd conv. layer & \multicolumn{1}{l|}{}                                     & \multicolumn{1}{l|}{}                                     & \multicolumn{1}{l|}{}                                       & 64 $\times$ 128 $\times$ 1 $\times$ 1  \\ \hline
		ReLU            & \multicolumn{1}{l|}{}                                     & \multicolumn{1}{l|}{}                                     & \multicolumn{1}{l|}{}                                       & \multicolumn{1}{c|}{\checkmark}                  \\ \hline
		4th conv. layer & \multicolumn{1}{l|}{}                                     & \multicolumn{1}{l|}{}                                     & \multicolumn{1}{l|}{}                                       & 3 $\times$ 64 $\times$ 1 $\times$ 1   \\ \hline
	\end{tabular}
	}
\end{table}

\begin{table}[h!]
\color{black}
\caption{Average Precision of MTDA-DTM with different architecture of DTM on Cityscape scenario.}
\label{tb:dtm_cs_arch_result}
\centering
\resizebox{0.8\columnwidth}{!}{
\begin{tabular}{|l||r|r|r|r|r|r|r|r||r|}
\hline
\textbf{Backbone}: VGG16  & \multicolumn{9}{c|}{\textbf{Accuracy}}           \\ 
\textbf{DTM's architecture}          & \textbf{bus}  & \textbf{bicycle} & \textbf{car}   & \textbf{m.cycle} & \textbf{person} & \textbf{rider} & \textbf{train} & \textbf{truck} & \textbf{mAP}      \\ \hline \hline
\multicolumn{10}{|l|}{\textbf{Train: Cityscape (\textbf{$S$})  $\xrightarrow{}$ FoggyCityscape (\textbf{$T_1$})  $\xrightarrow{}$ RainCityscape (\textbf{$T_2$}) - Test: all targets}}                                                         \\ \hline \hline
Source Only                        & 44.0          & 26.6          & 34.3          & 10.9          & 23.9          & 42.4          & 16.5          & 11.5          & 26.3          \\ \hline \hline
	UFT                                 & 65.5 & 34.3          & 48.1          & 18.7          & 28.0          & 53.3          & 35.5 & 25.9          & 38.6 \\ \hline
	Incr. MTDA KD \cite{wei2020_inc_mtda} & 61.9          & 34.2          & 48.5 & 21.2          & 28.8          & 52.8          & 32.8          & 24.6          & 38.1          \\ \hline
Original DTM        & \textbf{66.0}   & 34.6   & 48.4 & 18.2       & 29.3  & 53.5  & 35.4 & \textbf{27.6} & \textbf{39.1} \\ \hline \hline
Wider DTM       & 65.2 & 34.0    & 47.6  & \textbf{21.8}       & 28.2   & \textbf{56.4}  & 35.6  & 25.5  & 38.9  \\ \hline
4 Layers DTM       & 64.7 & 34.8    & 48.2  & 21.5         & 29.3   & 53.6  & \textbf{36.1}  & 26.7  & \textbf{39.4}  \\ \hline
6 Layers DTM       & 60.3 & \textbf{35.8}    & \textbf{48.7}  & \textbf{21.7}       & \textbf{29.5}   & 53.9  & 33.3  & 26.9  & 38.8 \\ \hline
\end{tabular}
}
\end{table}

\textcolor{black}{
From Table \ref{tb:dtm_cs_arch_result}, we observe that varying DTM's architecture only results in a slight change in performance. In this experiment, using a wider DTM lead to slight decreases  performance, whereas a deeper DTM like the "4 Layers DTM" can lead to a slight increase in results. Observing the results of "6 Layers DTM", we can see a decrease in performance of 3\%, which indicate the presence of a diminish return problem where a deeper DTM can impact performance negatively. In spite of this, our experiment shows that the performance of DTM is stable when there are small variations in the DTM architecture. 
}

\section{\textcolor{black}{UMAP of features of Source, DTM-transferred and Targets}}

\begin{figure}[htbp]
    \centering
\includegraphics[width=0.6\textwidth]{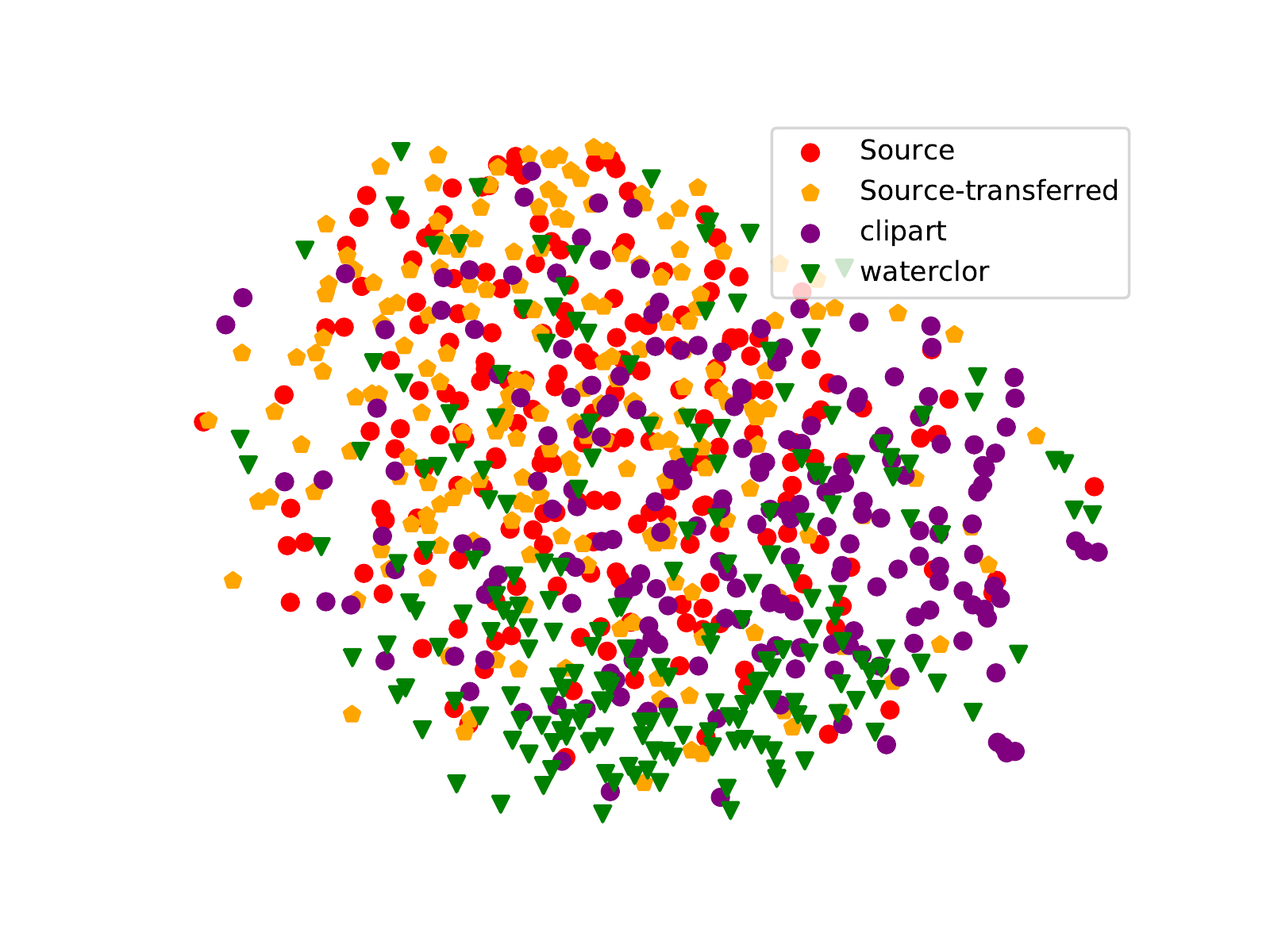}
    \caption{UMAP feature visualization of source domain (red), domain-transferred source domain (orange), all target domains: Clipart (purple) and Watercolor (green) from the DTM trained for scenario PascalVOC (\textbf{$S$})  $\xrightarrow{}$ Clipart (\textbf{$T_1$})  $\xrightarrow{}$ Watercolor (\textbf{$T_2$})  $\xrightarrow{}$ Comic (\textbf{$T_3$}) . This figure shows the desired outcome of our DTM. Best viewed in color.}
    \label{fig:umap_mask_vs_source_vs_targets_with_source}
    \vspace{-2mm}
\end{figure}

\textcolor{black}{For ease of visualization, we add a separate figure for UMAP visualization of source, DTM-transferred, and target features.}
\textcolor{black}{Figure \ref{fig:umap_mask_vs_source_vs_targets_with_source} shows that DTM features are slightly offset compared to source features. DTM does tend to shift the source features but, as shown in Figure \ref{fig:boundary}, this shift is small. In addition, our loss in Eq. \ref{eq:dtm_loss} does not encourage a large shift} \textcolor{black}{since we do not minimize the probabilities of classifying DTM features from the source to be zero. Rather, it encourages DTM features to remain consistent with the common source-target features while pushing the features closer to target domains. Additionally, since DTM samples are based on the same principle as adversarial examples, specifically targeting features, thus these will not show a large shift compared to the source}. \textcolor{black}{Finally, the visualization of these features from a trained models is due to the domain adaptation method, that encourages domain confusion in a model's feature representation to overcome domain shift (as seen in the supplementary material of \cite{HTCN}).} 

\end{document}